\RequirePackage{fix-cm}
\documentclass[10pt,a4paper]{article}
\usepackage[utf8]{inputenc}
\usepackage[T1]{fontenc}
\usepackage[table]{xcolor}
\usepackage{amsmath}
\usepackage{newtxmath}\usepackage{XCharter}\usepackage[scale=0.95]{sourcesanspro}
\usepackage{graphicx,array,float,booktabs,multirow,xspace,microtype}
\usepackage{algorithm,algorithmic,adjustbox}
\usepackage[section]{placeins}
\usepackage{flafter} % Keep figures after their source anchor.
\usepackage[a4paper,left=25mm,right=25mm,top=24mm,bottom=24mm,headheight=16pt,headsep=15pt,footskip=26pt]{geometry}
\usepackage{titlesec,fancyhdr,caption,tikz}
\usepackage[most]{tcolorbox}
\usepackage{cite}
\PassOptionsToPackage{hyphens}{url}
\usepackage{url,hyperref}
\definecolor{ink}{HTML}{18232E}
\definecolor{accent}{HTML}{245CF2}
\definecolor{pale}{HTML}{F0F4FC}
\definecolor{muted}{HTML}{5C6772}
\definecolor{tableodd}{HTML}{F0F4FC}
\definecolor{tableeven}{HTML}{F8FAFD}
\color{ink}
\hypersetup{colorlinks=true,linkcolor=accent,citecolor=accent,urlcolor=accent,
 pdftitle={System-One RSI: Building Muscle Memory for Real-Time Agents through Reasoning-Guided Few-Shot Practice},
 pdfauthor={Qingxu Fu, Xuanfa Jin, Ziqian Chen, Zhaoyang Liu, Bolin Ding}}
\newcommand{\rec}{System-One RSI\xspace}

\arrayrulecolor{black}
\makeatletter
\AtBeginEnvironment{table}{\rowcolors{1}{tableodd}{tableeven}}
\AtEndEnvironment{table}{\global\@rowcolorsfalse}
\AtBeginEnvironment{table*}{\rowcolors{1}{tableodd}{tableeven}}
\AtEndEnvironment{table*}{\global\@rowcolorsfalse}
\makeatother
\titleformat{\section}{\sffamily\Large\bfseries\color{ink}}{\colorbox{accent}{\color{white}\makebox[1.1em]{\thesection}}}{0.7em}{}
\titleformat{\subsection}{\sffamily\large\bfseries\color{ink}}{\color{accent}\thesubsection}{0.65em}{}
\titleformat{\subsubsection}{\sffamily\normalsize\bfseries\color{ink}}{\thesubsubsection}{0.6em}{}
\titlespacing*{\section}{0pt}{22pt plus 3pt minus 2pt}{9pt}
\titlespacing*{\subsection}{0pt}{15pt plus 2pt minus 1pt}{6pt}
\renewenvironment{abstract}{\begin{tcolorbox}[enhanced,breakable,colback=pale,colframe=pale,boxrule=0pt,sharp corners,left=12pt,right=12pt,top=9pt,bottom=9pt]\sffamily\bfseries\small\color{accent}ABSTRACT\par\vspace{4pt}\normalfont\fontsize{10}{13}\selectfont\color{ink}}{\end{tcolorbox}}

\makeatletter
\renewcommand{\maketitle}{%
  \thispagestyle{empty}
  {\sffamily\fontsize{9}{11}\selectfont\color{muted}PREPRINT\hfill\MakeUppercase{\@date}\par}
  \vspace{15pt}
  {\sffamily\fontsize{38}{40}\selectfont\bfseries\color{ink}\papertitleshort\par}
  \vspace{10pt}
  {\sffamily\fontsize{21}{25}\selectfont\color{ink}\papersubtitle\par}
  \vspace{10pt}
  {\color{accent}\rule{52mm}{3pt}\par}
  \vspace{14pt}
  \@author
  \vspace{12pt}
}
\makeatother

\newcommand{\papertitleshort}{System-One RSI}
\newcommand{\papersubtitle}{Building Muscle Memory for Real-Time Agents\\through Reasoning-Guided Few-Shot Practice}
\title{System-One RSI: Building Muscle Memory for Real-Time Agents through Reasoning-Guided Few-Shot Practice}
\date{November 2026}

\newcommand{\paperauthors}{%
{\sffamily\fontsize{10}{13}\selectfont\bfseries Qingxu Fu$^{1}$\quad Xuanfa Jin$^{1,2}$\quad Ziqian Chen$^{1}$\quad Zhaoyang Liu$^{1}$\quad Bolin Ding$^{1}$\par}
\vspace{5pt}
{\sffamily\fontsize{8.5}{11}\selectfont
$^{1}$Alibaba Token Foundry - Infra \& Model RSI - EconML\\
$^{2}$Institute of Automation, Chinese Academy of Sciences\par}
\vspace{5pt}
{\sffamily\fontsize{8}{10.5}\selectfont\color{muted}
\texttt{qingxu.fu@outlook.com}\quad\texttt{jinxuanfa.jxf@alibaba-inc.com}\\
\texttt{eric.czq@alibaba-inc.com}\quad\texttt{jingmu.lzy@alibaba-inc.com}\quad\texttt{bolin.ding@alibaba-inc.com}\par}}
\author{\paperauthors}

\begin{document}
\maketitle

\begin{abstract}
% // AI generated
% AI Summary: 本文借鉴人类在指导下练习形成肌肉记忆的过程，从少量真实交互中获得监督，递归改进 JEV 的实时视觉判断。
Reasoning models excel at following complex instructions and understanding task logic, but incur high latency during deliberation, whereas decision models respond quickly yet can struggle with instruction following and complex task logic.
We present System-One Recursive Self-Improvement (System-One RSI), which couples an online Observe, Orient, Decide, Act (OODA) loop with an offline improvement loop.
During live execution, the real-time agent follows the OODA cycle: a decision model determines task-relevant visual features, which are used by a rule-based controller to select actions.
To achieve policy optimization in an outer RSI loop, the reasoning agents analyze interaction trajectories, build datasets,
and fine-tune the decision model parameters or simply adjust rules and prompts.
Through recursive few-shot practice, 
the experience gained from the reasoning agents as well as environment interaction is consolidated into the real-time agent as decision model parameters.
This framework resembles the acquisition of muscle memory and enables learning from a small number of sequential real-world episodes without expert action demonstrations, sandboxing, or parallel execution, overcoming the limitations of many online RL (reinforcement learning, requiring environment sandboxing) and VLA (vision language action, requiring expert data) methods.
Experiments use a 27B decision model in the game Snake, Zenless Zone Zero (ZZZ), and Black Myth: Wukong (BMW).
The main experiments are performed with the controller fixed.
In the game Snake, the mean score rises from 1.30 to 7.35;
in an evaluated ZZZ scenario, the attack-warning accuracy increases from 44.52\% to 77.98\%, 
and the median completion time among winning episodes decreases by 8.4\%;
in a BMW combat level, the evasive policy extends the average survival time from 108.1 to 260.5 seconds;
and the offensive policy raises the win rate from 0\% to 66.7\%.

\end{abstract}

\clearpage

\section{Introduction}
\label{sec:intro}

\begin{figure}[!htbp]
\centering
\setlength{\tabcolsep}{2pt}
\begin{tabular}{@{}cccc@{}}
\includegraphics[width=\dimexpr(\linewidth-6\tabcolsep)/4\relax,trim=187.2bp 0bp 187.2bp 0bp,clip]{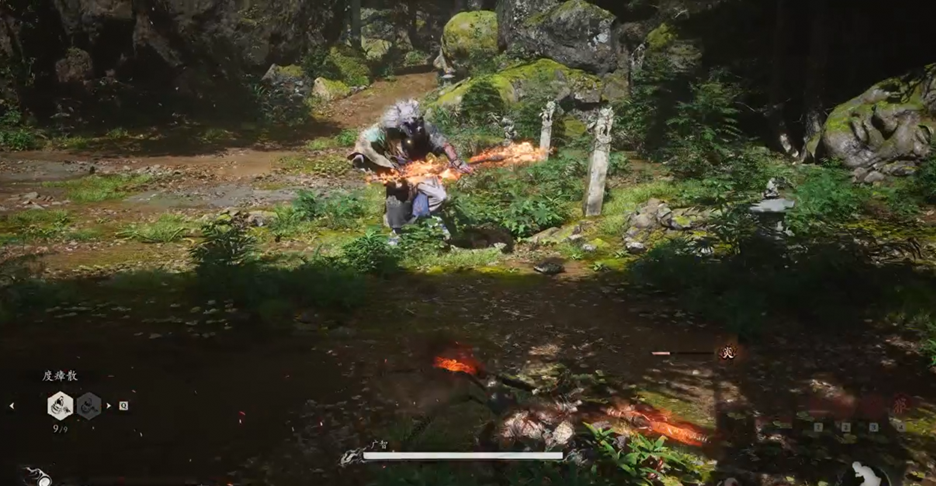} &
\includegraphics[width=\dimexpr(\linewidth-6\tabcolsep)/4\relax,trim=187.2bp 0bp 187.2bp 0bp,clip]{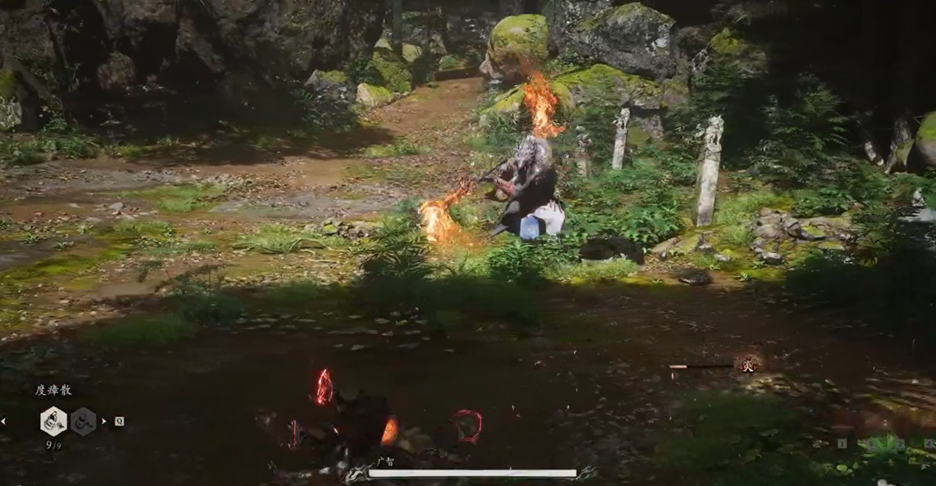} &
\includegraphics[width=\dimexpr(\linewidth-6\tabcolsep)/4\relax,trim=187.2bp 0bp 187.2bp 0bp,clip]{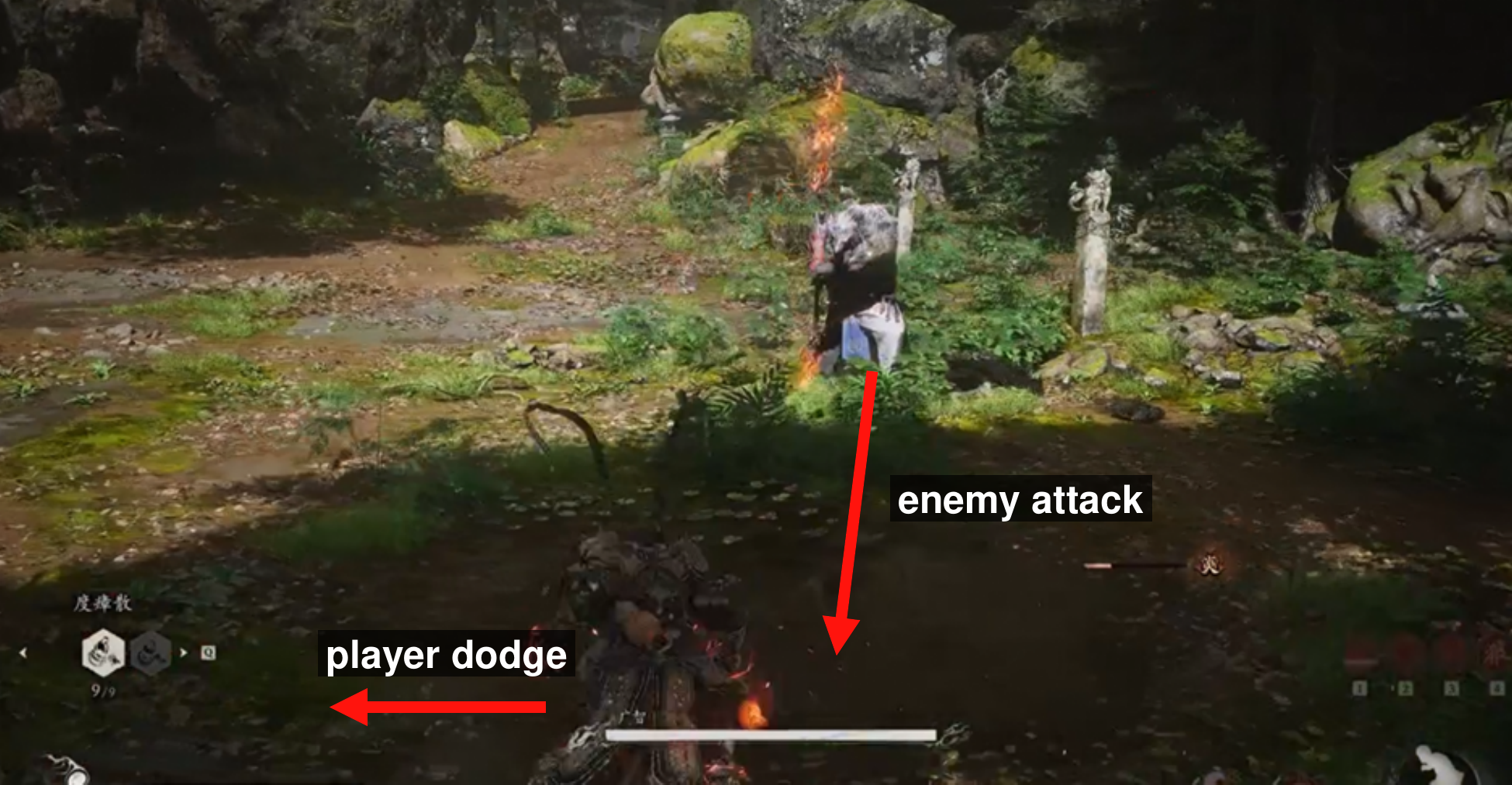} &
\includegraphics[width=\dimexpr(\linewidth-6\tabcolsep)/4\relax,trim=187.2bp 0bp 187.2bp 0bp,clip]{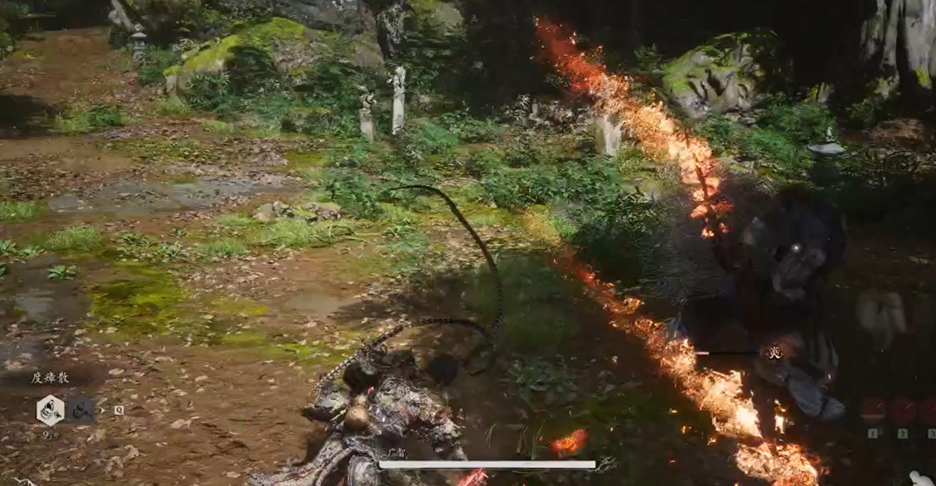} \\
(a) Pre-attack & (b) Pre-attack & (c) Dodge window & (d) After attack
\end{tabular}

% AI generated
\caption{Successive observations of an attack in \emph{Black Myth: Wukong}. Panels (b) and (c) both show the opponent with a raised weapon. The dodge window opens in (c), early action in (b) will cause failure. For static decision models, text prompting cannot easily distinguish (b) and (c), and few-shot images cause latency impact.}
\label{fig:prompt-case}
\end{figure}

% Human - AI Assisted
% Preserve the human baseline: reasoning-enabled task competence, computer use,
% robotics, reasoning-effort cost, model-depth cost, and sub-second control.
%%%%%%%
% Human Baseline: 受益于长链推理能力，大语言模型不仅能解决复杂的文本处理任务，还能完成出色的多模态环境交互任务，如 Computer Use 等，作为大脑控制机器人等。
% 虽然借助推理能力的模型在几乎所有维度上都取得了显著的提升，但推理链条（reasoning effort）开销越大，决策花费的时间也跟着增长。此外，层数越多的模型单个token推理的时间和内存开销越大。
% 综合这些因素，推理大语言模型很难实现真实的亚秒级控制任务……
%%%%%%%
% Human Summary: 推理能力让语言模型能处理多模态交互任务，但推理链越长、模型越深，单次决策越慢，因此推理模型难以在亚秒级窗口内完成控制。
Long-chain reasoning enables language models to develop and revise intermediate steps before producing an answer, helping them decompose complex tasks and use intermediate results to guide subsequent decisions~\cite{yu2024system2}.
Combined with multimodal perception and action interfaces, these capabilities enable complex software engineering, desktop interaction, and high-level planning for robots~\cite{yang2024sweagent,yuan2026osworld,ahn2022saycan}.
However, greater reasoning effort requires more intermediate tokens, each adding a sequential decoding step.
In a continuously changing environment that requires low-latency responses (such as Fig.~\ref{fig:prompt-case}), 
this latency can leave observations stale or 
cause an action opportunity to pass before the reasoning that identifies it has finished~\cite{tang2026realtimebench}.

% Human - AI Assisted
% Preserve the human baseline: decision readouts, single-pass latency, the loss
% of explicit multistep deliberation, prompt/input costs, and learning from practice.
%%%%%%%
% Human Baseline: 近期决策模型（decision model）重新收到越来越多的关注，这类决策模型通常使用和普通大语言模型相似的基础结构，最大的区别在于输出层。[描述下decision输出层的差别]……。其次，给定输入后，决策模型只进行单次输出，再加上层数不多，因此多数决策模型，能够将决策时间压缩到 50 至 200 毫秒的区间内，足够满足绝大多数实时系统的需求。
% 但遗憾的是，脱离推理能力的模型无法像一般语言模型一样将决策过程分散到多个步骤中，这导致做出正确决策的难度陡增。如果在具体问题下进行细致的提示词微调，可以达到较高精度，但较长提示词或多幅图像输入容易推理耗时增加至原来的若干倍，耗时会退化到和一般语言模型相似的水平。
% 此外，有些决策场景（如[我们下文给出的4图判断躲避实际]）在理论上也完全无法使用少量的提示词和few-shot参考图实现决策，因为智能体必须在练习和交互中获取经验，并充分利用这些经验才能做出正确的决策。
%%%%%%%

Decision models such as JEV reduce this decoding overhead by scoring candidate answers directly~\cite{typesafe2026jev,li2026llmasjev}.
They retain language-model backbones but use a decision head or selected output logits to produce candidate scores~\cite{zhang2026medjev}.
However, these models lack explicit intermediate reasoning steps for decomposing and revising decisions and are unable to perform deep analysis for a complex decision-making process.
Task-specific prompt tuning may improve accuracy, but longer prompts and additional reference images can erode the latency advantage.
Moreover, in some cases prompts (including text with several reference images) cannot convey the fine-grained multimodal experience.
For example, in the game Black Myth: Wukong, dodge timing requires distinguishing the visually similar postures in Fig.~\ref{fig:prompt-case}(a)--(c) and initiating a dodge at (c).
Both (b) and (c) can be described as an opponent raising a weapon, yet this description does not distinguish their action timing.
When prompting alone cannot supply this distinction, interaction sequences and their outcomes provide supervision for updating the decision model's weights, consolidating experience from practice into features that can be predicted rapidly during subsequent execution.

% Human - AI Assisted
% 段落要点 承接上一段的经验获取问题，用强化学习说明策略可以从交互中获得，而训练后的浅层策略仍可快速执行。
% 写作思路 保留原稿对在线学习成本的解释。账号认证和服务端持久状态会影响隔离、重置和并行实例，串行采集时，每条新增轨迹都要消耗实际运行时间。
% 写作思路 离线学习可以复用已有轨迹，关键约束转为轨迹的采集成本，以及数据是否覆盖相关状态、动作和后果。
% 衔接逻辑 第三段聚焦经验来源，表 1 用原稿的优势与局限列汇总各条路线，第四段再从对照中归纳推理模型与决策模型的互补作用。
Reinforcement learning (RL) offers a way to acquire task-specific control through experience.
Once trained, a shallow policy network can map an observation to an action in a short forward pass.
Obtaining that policy, however, can require substantial interaction data~\cite{berner2019dota,dulac2019challenges}.
Online reinforcement learning gathers experience through repeated trials and updates the policy using reward feedback.
In a live game client, dependencies such as account authentication and persistent service state can make it difficult to isolate the environment, reset it reliably, or run independent copies in parallel.
When trials must run one at a time, collecting another trajectory means running another trial at the environment's own pace.
Offline reinforcement learning can learn from existing trajectories without further interaction during training.
Those trajectories still have to be collected, and their coverage limits what can be learned about actions whose consequences are poorly represented in the data~\cite{levine2020offline}.
Table~\ref{tab:approaches} summarizes these approaches alongside vision-language-action (VLA) models and decision models, comparing how they acquire control capability and the constraints on its use in real time.

\begin{table}[!htbp]
\centering
% AI generated
% 本表比较不同路线如何获得控制能力及其在目标环境中的主要限制。
% 汇总决策模型段、强化学习段与提示词/代码段的分析。
% 下一正文段转入人类学习类比，作为综合这些路线的出发点。
% 本表为方法选择提供相关工作背景，所列限制属于机制与部署条件的比较。
\caption{Routes to control capability and their constraints in live environments.}
\label{tab:approaches}
\small
\begin{tabular}{@{}>{\raggedright\arraybackslash}p{0.20\textwidth}>{\raggedright\arraybackslash}p{0.29\textwidth}>{\raggedright\arraybackslash}p{0.45\textwidth}@{}}
\toprule
Approach & Advantages & Limitations \\
\midrule
Online RL~\cite{berner2019dota,dulac2019challenges} & Learns policies from interaction feedback & Environment replication, reset, and parallel rollouts may be obstructed by external dependencies. \\
Offline RL~\cite{levine2020offline} & Learns from existing trajectories & Data collection can be costly; coverage limits learning under distribution shift. \\
Reasoning computer use~\cite{xie2024osworld,lin2023swiftsage} & Applies task knowledge through deliberation & Long reasoning adds latency, and visual grounding can remain unreliable. \\
VLA, demonstration-trained~\cite{chen2025combatvla} & Learns action sequences from gameplay videos & Requires hours of gameplay from experienced players and manual annotation and cross-validation of question-answer and action-of-thought data. \\
VLA, large-scale trajectory + online RL~\cite{luo2026gamer} & Learns from large-scale gameplay trajectories and online RL & Uses thousands of hours of gameplay data and an sandbox that modifies the game engine (such as Unity) for online RL. \\
Pretrained JEV~\cite{typesafe2026jev} & Returns structured features without a reasoning trace & Pretrained capabilities may not cover the distinctions required by a new task. \\
JEV prompt optimization~\cite{typesafe2026jev,liu2021prompt} & Revises instructions without parameter updates & Fine distinctions can be difficult to express; longer inputs increase processing cost. \\
\bottomrule
\end{tabular}
\end{table}
\FloatBarrier

\begin{figure}[!htbp]
\centering
\includegraphics[width=\textwidth]{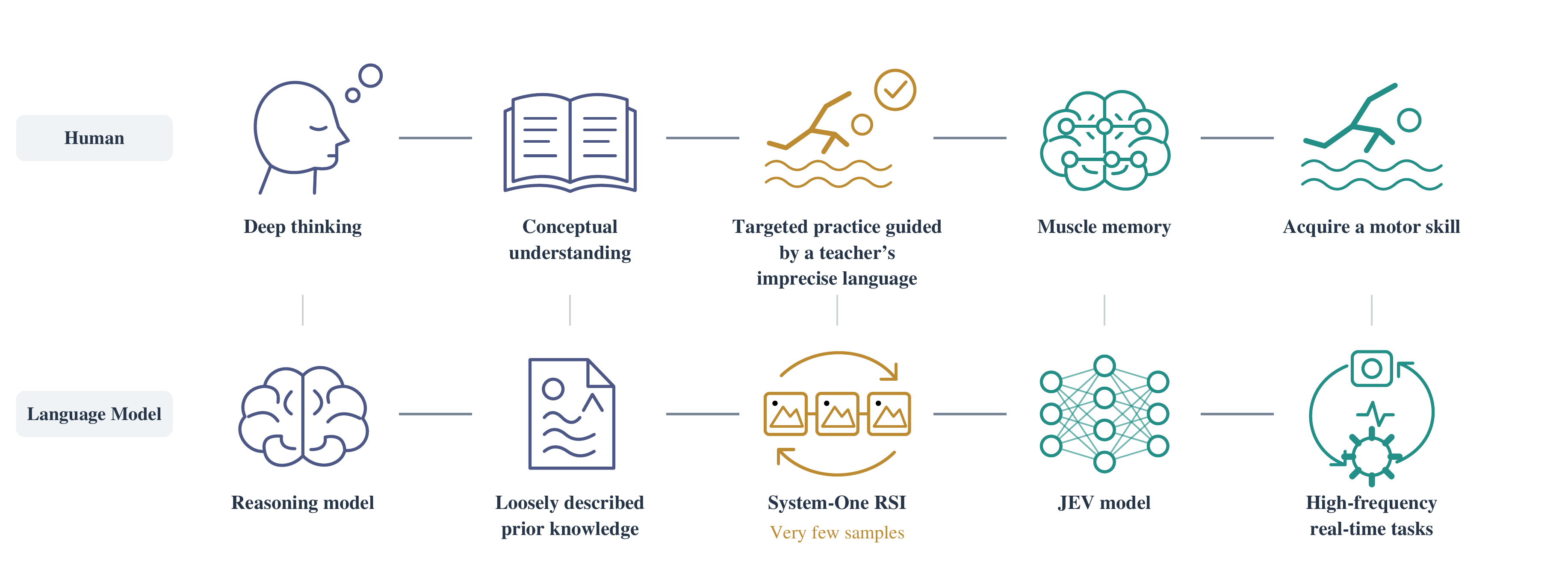}
\caption{Guided practice as a model for learning external muscle memory.}
\label{fig:rsi-concept}
\end{figure}

% Human - AI Assisted
% 段落要点 保留表后归纳的三步。先把推理模型的任务理解和决策模型的快速判断联系起来，再限定问题为只用少量串行实战、环境保持原状且没有专家操作示范时如何学习。
% 写作思路 用游泳与图 2 对应。学习者理解呼吸与划水的配合，仍要靠练习掌握时机。教练观察哪里失去节奏，并据此指导下一次练习。
% 写作思路 把类比继续讲到动作趋于自动化，再映射到模型。推理模型从尝试中确定要学的内容，决策模型参数中可快速调用的任务判断构成外置肌肉记忆。
% 衔接逻辑 先说明重新部署会改变智能体的动作和后续反馈，再用过渡句和问句把两个模型的分工收束到 RSI。第五段直接用框架结构回答这个问题。
The comparison in Table~\ref{tab:approaches} highlights two capabilities that can work together.
A reasoning model can draw on prior knowledge to understand an unfamiliar task, and a decision model can provide fast features during execution.
The challenge is to acquire those features from a small number of episodes collected one at a time in an unmodified environment, without expert action demonstrations.
A person learning to swim faces a similar problem.
They may know how breathing and strokes should be coordinated, yet getting the timing right in the water takes practice.
A coach can watch a few strokes, notice when the learner loses rhythm, and use that observation to guide the next exercise~\cite{ericsson1993practice}.
With practice, the swimmer can make these adjustments without consciously working through each movement, a capacity commonly described as muscle memory.
Figure~\ref{fig:rsi-concept} shows how this idea applies to an agent.
A reasoning model reviews the agent's attempts and specifies what the decision model should learn from them.
The task-specific features retained in the decision model's parameters and available for rapid use are called \emph{external muscle memory}.
When the updated decision model returns to the environment, its features shape the actions the agent takes and the feedback it receives.
Combining guidance from the reasoning model with fast execution by the decision model raises a natural question:
Can the reasoning model use this feedback to refine the decision model over successive rounds of recursive self-improvement (RSI), enabling the agent to act more reliably within the time available?

\begin{figure}[!htbp]
\centering
\includegraphics[width=\linewidth]{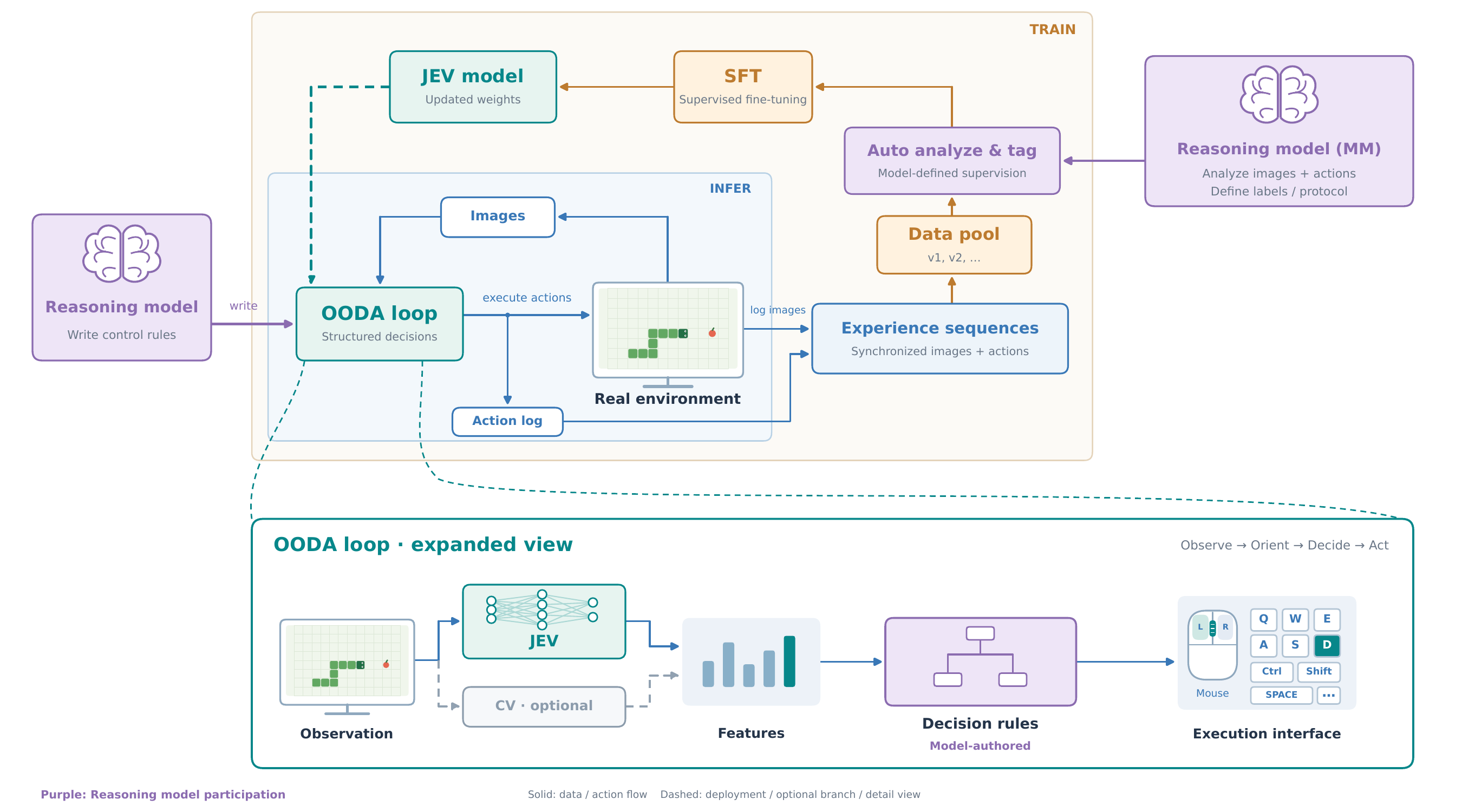}
% Relocated from the opening of the original method chapter.
\caption{Online control and recursive improvement in \rec{}.}
\label{fig:framework}
\end{figure}

% Human - AI Assisted
% 段落要点 按原稿介绍双环的顺序展开框架。正文首次出现时给出 OODA 全称，再说明模型按任务问题输出结构化判断，决策树据此选动作和时机，执行接口提交动作。
% 写作思路 按一轮实践介绍外环。先定义任务问题和动作规则，再运行当前策略采集记录，最后由推理智能体回看观测、判断和动作结果。
% 写作思路 吸收新版方法对失败来源的区分。可见线索已经进入模型但被误判时，可以构造监督并微调 JEV。其他失败则可据记录调整视觉输入、任务问题或控制规则。
% 衔接逻辑 以评估和重新部署结束这一轮，新系统再产生下一轮经验。这把第四段的递归问题落实为可执行的流程，为下一段的实战结果做铺垫。
This paper presents \rec{}, which couples a fast online decision loop with an offline improvement loop (Fig.~\ref{fig:framework}).
The inner loop follows the Observe, Orient, Decide, Act (OODA) cycle.
At each step, the decision model receives a visual observation and returns a structured feature state for each task-relevant question.
A rule-based controller combines these feature states to select an action and determine its timing, and the execution interface sends that action to the environment.
This loop runs without consulting a reasoning model during execution.
The outer loop is coordinated by a team of reasoning agents.
They define the questions the decision model should answer and write the rules that map those answers to actions.
They then deploy the controller and collect interaction records, starting from a simple policy when needed.
Between rollouts, the reasoning agents review what the system observed, how JEV interpreted it, and what followed the action.
When a relevant cue was available to JEV but its feature state was wrong, the agents use the recorded evidence to define supervision and fine-tune the model.
The same records can guide revisions to the visual input, feature questions, or controller rules when the failure originates in one of those components.
After evaluation and redeployment, the revised system generates the experience used in the next round.

% Human - AI Assisted
% 段落要点 按各实验能说明什么组织结果。Snake 的总体得分提高，中间回退说明每次更新都值得单独评估。悟空同配置下的闪避时长说明时机判断可以支持持续闪避，另报进攻配置的任务完成表现。
% 写作思路 绝区零先报验证集上的平衡准确率，再用另一组固定控制器的在线对照，检验模型微调能否在决策耗时相近的情况下改善任务完成表现。
% 写作思路 将胜局完成时间和决策轮次耗时分开。前者只统计完成的回合，后者约 230 毫秒，口径是从采帧到选定动作。
% 衔接逻辑 引言保留支撑结论的核心结果和关键对照条件。模型规模、采集轮数、验证帧数等设置由后续章节交代。
The evaluation covers three dynamic visual environments.
In Snake, the mean score rises from 1.30 to 7.35, with a decline in an intermediate version that illustrates the importance of evaluating each update.
In \emph{Zenless Zone Zero}, balanced accuracy for attack-warning features rises from 44.52\% to 77.98\% on the common 260-frame v2 validation set.
In a separate online comparison with the controller held fixed, both fine-tuned models complete five of five episodes, compared with two of five for the reference.
For the later checkpoint, median completion time among wins falls from 261.2\,s to 239.3\,s, while the median decision round from frame capture to action selection remains near 230\,ms.
In these runs, fine-tuning improved task completion at a similar decision-round duration.
In \emph{Black Myth: Wukong}, mean recorded evasion duration rises from 160.9\,s to 260.5\,s under the same controller configuration, illustrating how learned timing features can support sustained evasion.
A separate offensive configuration wins 10 of 15 episodes, showing how those features can be used alongside attack and healing rules to complete the task.

% 段落要点 说明后文如何展开。方法章解释在线控制与递归改进，系统章介绍多智能体实现和帧处理，实验章报告视觉判断和实战结果。
% 写作思路 导读反映新版论文独立设置 System 章的结构。模型规模与视觉输入的比较在该章给出时限和测量方案，当前表格仍待补测，因此正文写比较方法。
Section~\ref{sec:method} presents the online control loop and the procedure for recursively improving JEV.
Section~\ref{sec:system} describes the multi-agent implementation and online frame processing, and defines how to compare visual inputs and model sizes under a reaction deadline.
Section~\ref{sec:results} reports the evaluation of visual features and live task outcomes.

\section{Method}
\label{sec:method}

% // AI generated
This paper presents System-One RSI for tasks in live environments with streaming visual observations and an executable action interface.
Timeliness is critical: the online loop must act promptly before the observations become outdated.
The improvement loop, in turn, encloses the inner loop, identifies its failures, and determines whether the next update should tune the decision model, revise the rule-based controller, or update both.

\subsection{OODA with decision model}
\label{sec:student}
\label{sec:controller}
The inner loop is an OODA cycle built around a decision model and a rule-based controller. As shown in Fig.~\ref{fig:framework}, it has two steps: (1) determine the feature states of the current environment, and (2) produce the final action.

At each decision step $i$,
the latest frames are captured from the environment.
The observation is denoted as $o_i=\mathcal{W}(I_i,H_i)$,
where $I_i$ represents the latest environment image, $H_i$ represents historical images captured shortly before $I_i$, and $\mathcal{W}$ reduces their resolution.
The chosen input representation determines which spatial and temporal evidence reaches the decision model.

To determine the feature states of the current environment,
the feature question $q_k \in \mathcal{Q}$ is supplied to the decision model as a prompt together with observation $o_i$.
For each question,
the decision model $f_\theta$ estimates a probability distribution over the possible states of each feature dimension and greedily selects one state for each feature.
For the $k$-th feature, the state is computed as
\begin{equation}
\begin{aligned}
s_{ik} &= \arg\max_{s \in S_k} f_\theta(s \mid o_i,q_k),\\
\end{aligned}
\label{eq:online-control}
\end{equation}
where $s_{ik}$ is the feature state selected at step $i$.
$S_k$ is the finite set of possible states for the $k$-th feature.
The vector $\boldsymbol{s}_i = \lbrace s_{i1}, s_{i2}, \dots \rbrace$ collects the states of all features.

For simple games such as Snake,
a feature can be defined directly as the output action, such as one of four two-dimensional directions.
This approach is difficult to apply to games with more sophisticated reaction logic.
An additional rule-based controller $C$ is therefore needed to integrate state information and select actions.

The controller rules $\phi$ are generated by the outer-loop reasoning model.
The controller maintains internal states $e_i$.
These states support complex logic such as combo actions and can filter incorrect feature states using temporal information, since the decision model is stateless and not always accurate.

\begin{equation}
\begin{aligned}
\boldsymbol{a}_i &= C(\boldsymbol{s}_i, e_i \mid \phi).
\end{aligned}
\label{eq:online-control-2}
\end{equation}
where $\boldsymbol{a}_i$ denotes the action at step $i$.
The action is multidimensional: it specifies (1) which keys should be pressed, (2) how long each key should remain pressed, and (3) where the mouse should be positioned when a mouse action is involved.

\subsection{Recursive improvement in \rec{}}
\label{sec:recursive-improvement}
Real-time response depends on the quality of the features that arrive within the available response window.
Nevertheless, the fast OODA loop built around the System-One model may produce unreliable features because
(1) the decision model's prompt is not properly tuned,
(2) the decision model's capability is limited and requires weight tuning,
(3) the rule-based controller encounters unexpected situations not covered by its rules.
A recursive improvement loop is therefore essential for continuous online policy improvement.
The online loop is treated as part of the RSI cycle,
providing the trajectories that record all previous environment interactions.
Figure~\ref{fig:framework} shows how a rollout supplies experience for improvement and how the resulting model returns to online control.

\textbf{OODA initialization}.
When starting from scratch,
the only available materials about a specific environment are assumed to be (1) a vague human-provided prompt describing basic knowledge, key mappings, and potential policies and, optionally, (2) public materials obtained through search tools, including web videos and blogs.
The reasoning model creates the first version of the OODA online cycle, including the decision-model questions and feature space, as well as the rules for the controller.
Because accuracy depends on the vague prior-knowledge prompt and the capability of the reasoning model, the initial OODA loop is usually unreliable. A dedicated initialization step therefore calibrates its inputs and outputs, ensuring that observation images meet the resolution and latency requirements and that the key mapping is correct.

\textbf{Online Sampling}.
The system applies the OODA loop and runs episodes in the environment.
Because the environment is assumed not to support parallel execution,
only a small number of samples are collected sequentially as a batch
and then stored in a data pool.

The samples in the data pool are used to identify policy weaknesses or construct supervised-learning datasets.

\textbf{Diagnosis and choice of practice}.
As shown in Fig.~\ref{fig:multiagent},
an improvement cycle can follow one of two branches: (1) if feature-state estimation is inaccurate and the decision-model weights can be trained with supervision, choose \textit{A: decision model optimization cycle}; or (2) if the environment reveals new patterns that the reasoning model has not considered, choose \textit{B: prompt and controller optimization cycle}.

\textbf{A: Decision model optimization cycle}.
To reduce the influence of other variables, the controller rules and decision-model prompts remain fixed during decision-model optimization.
There are five steps:
(1) discover the features that suffer low accuracy;
(2) extract historical episode trajectories from the data pool;
(3) annotate the samples with reasoning models that have strong multimodal visual capability, and split the annotated dataset into a training set and a validation set;
(4) train the decision model on the accumulated datasets from step (3);
(5) deploy the new decision model and enter the next improvement cycle.

\textbf{B: Prompt and controller optimization cycle}.
This branch requires analyzing the entire episode video and investigating the environment's responses to the actions.
However, many reasoning models have strong image-understanding capability yet are not trained with video encoders and cannot analyze videos in a straightforward way.
To address this problem, we adopt an alternative approach:
instead of encoding all images into a video,
the agent responsible for episode analysis converts the video into low-resolution image montages resembling a film strip.
Inspecting montages taken at different moments and time intervals makes it easier to identify actions that do not work as expected and fix the underlying controller problem.

If the problem is traced to poor decision-model precision,
prompt tuning can be applied in two ways: revising the text prompt and adding reference images.

% \subsubsection*{JEV fine-tuning and redeployment}
% % // AI generated
% Training uses the selected examples to update the JEV parameters responsible for visual features. A candidate checkpoint is evaluated before redeployment in the online loop. Its features can change the actions executed and the states visited, giving the next round different evidence about what the system needs to learn.

\FloatBarrier
\subsection{Learning from few-shot episodes}
\label{sec:teacher}
\label{sec:pool}
The decision model is the core of the high-frequency decision loop.
The main challenge in supervised training of the decision model is that episodes are few and the available samples are limited and imbalanced.
To address this challenge, the method (1) establishes a data pool $\mathcal D$ that gathers samples across RSI iterations and (2) uses a focal loss function to balance their contributions.

\textbf{Data pool accumulation}. Let $r\in\lbrace 0,1,2,\dots\rbrace$ index the RSI iterations. Initialize the data pool as $\mathcal D(-1)=\varnothing$, and let $\mathcal T(r)$ denote the trajectories collected at iteration $r$. The weighted sampler produces training and validation subsets, denoted by $\mathcal D_{\mathrm{train}}^{s}(r)$ and $\mathcal D_{\mathrm{val}}^{s}(r)$, respectively:
\begin{equation}
\begin{aligned}
\mathcal T(r) &= \operatorname{Rollout}\left[\theta(r), \phi(r), \text{prompt}(r)\right],\\
\mathcal D(r) &= \mathcal D(r-1)\cup\mathcal T(r),\\
\mathcal D^{s}(r) &= \operatorname{WeightedSampler}\left[\mathcal D(r)\right].
\end{aligned}
\label{eq:agent-supervision}
\end{equation}
Here, $\operatorname{WeightedSampler}$ balances samples from different RSI iterations and partitions the sampled data into training and validation subsets.

\textbf{Loss function}. Let $\theta=(\bar\theta,\omega)$ denote the model parameters, where $\omega$ is the selected trainable subset and $\bar\theta$ denotes the fixed parameters. For a labeled training example comprising observation $o_i$, feature question $q_k$, and reference state $s\in S_k$, define the per-example focal loss and the iteration-$r$ training objective as follows:
\begin{equation}
\begin{aligned}
\alpha_{k,s} &= \frac{\rho_k(s)^{-1}}{\sum_{s'\in S_k}\rho_k(s')^{-1}},\\
\ell_{ik}(s;\theta) &= -\alpha_{k,s}\left(1-f_\theta(s\mid o_i,q_k)\right)^{\gamma}
\log f_\theta(s\mid o_i,q_k),\\
\mathcal L_r(\omega) &= \frac{1}{\left|\mathcal D_{\mathrm{train}}^{s}(r)\right|}\sum_{(o_i,q_k,s)\in\mathcal D_{\mathrm{train}}^{s}(r)}\ell_{ik}(s;\theta).
\end{aligned}
\label{eq:focal-loss}
\end{equation}
Here, $i$ indexes the observation, $k$ indexes the feature, $\rho_k(s)$ is the empirical frequency of state $s$ among training examples for feature $q_k$, and $\gamma=2$. The normalized inverse-frequency weights $\alpha_{k,s}$ are precomputed by the dataset builder and used unchanged during training. The class weights increase the relative contribution of rare states, while the focal factor down-weights well-classified examples.

\textbf{Rollout with new decision model}. Section~\ref{sec:results} gives the trainable subset and training choices used in the experiments. At iteration $r$, validation selects a checkpoint whose trainable parameters are denoted by $\omega(r+1)$. A model-only update deploys $\theta(r+1)=(\bar\theta,\omega(r+1))$ while holding $\text{prompt}(r)$ and $\phi(r)$ fixed. The next rollout, $\mathcal T(r+1)=\operatorname{Rollout}[\theta(r+1),\phi(r),\text{prompt}(r)]$, adds trajectories to the data pool in Eq.~\ref{eq:agent-supervision}.

\section{Implementation}
\label{sec:system}

\subsection{Multi-agent framework for RSI}
\label{sec:coordination}

Modern reasoning models can address complex tasks. However, when asked to manage long-running workflows involving (1) trajectory capture, (2) visual annotation, (3) supervised fine-tuning with distributed data parallelism (DDP), (4) model deployment with data parallelism (DP), and (5) planning and version management, they often encounter context-window limits or fail to follow basic instructions.

\begin{figure}[!htbp]
\centering
\includegraphics[width=\linewidth]{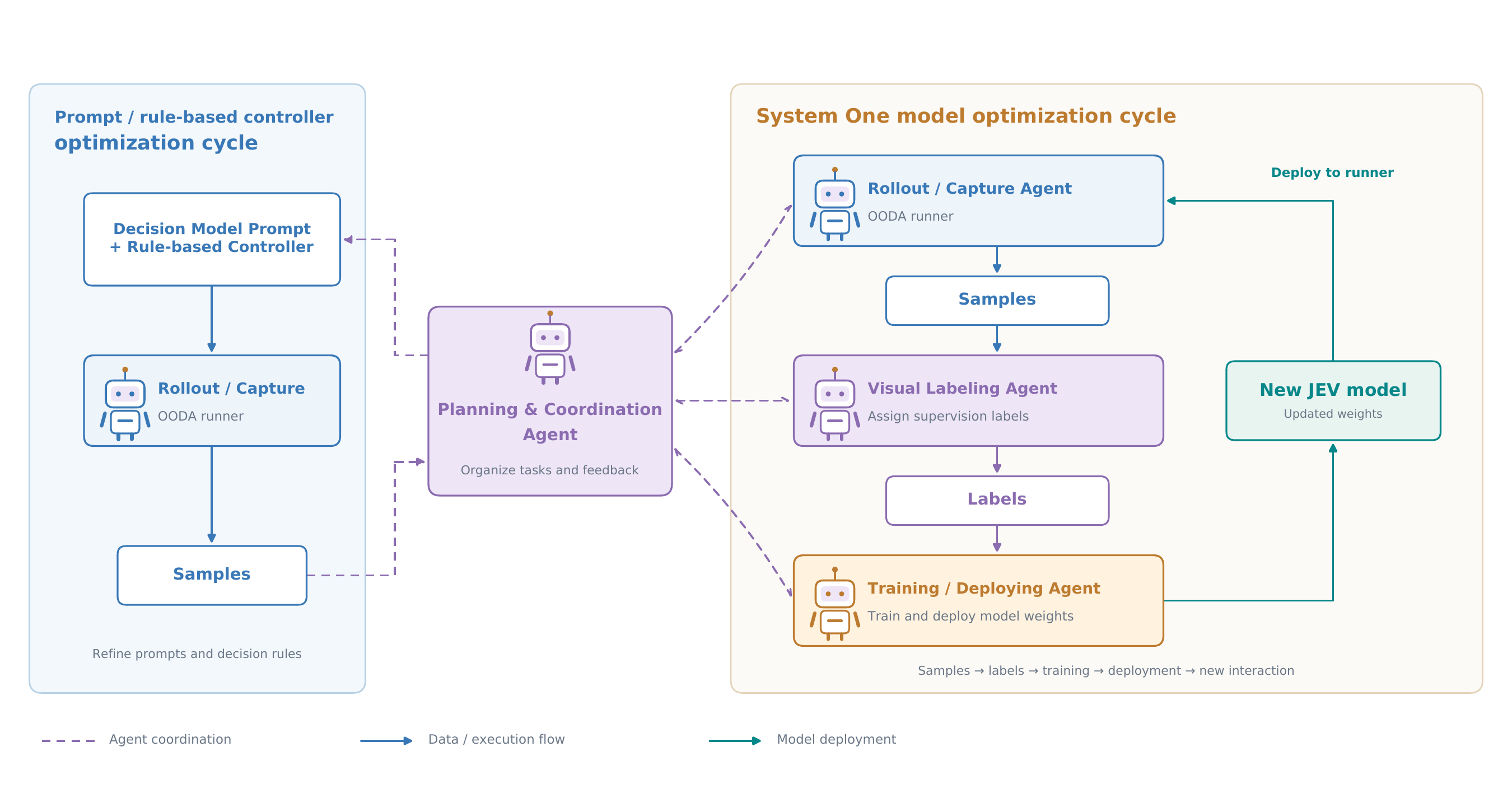}
\caption{Agent roles in controller revision and model fine-tuning.}
\label{fig:multiagent}
\end{figure}

The workflow is therefore divided among multiple agents with distinct instructions. As shown in Fig.~\ref{fig:multiagent}, four agents are assigned to the decision-model optimization workflow:

\begin{itemize}
  \item \textbf{Planning \& Coordination Agent.} Orchestrates capture, labeling, training, and deployment by assigning tasks and coordinating agent handoffs. It tracks progress, consolidates feedback, and manages model and dataset versions across iterations.
  \item \textbf{Rollout / Capture Agent.} Executes the OODA loop and records time-aligned observations, predicted feature states, actions, and outcomes. These trajectories supply annotation samples and support evaluation of the updated decision model.
  \item \textbf{Visual Labeling Agent.} Inspects recorded observations and their temporal context to assign reference states to feature questions. It passes the labeled examples to the Training / Deploying Agent for supervised learning and validation.
  \item \textbf{Training / Deploying Agent.} Prepares training and validation subsets from accumulated labeled examples and fine-tunes the decision model with DDP. It selects a checkpoint using validation performance and deploys it with DP for subsequent rollouts.
\end{itemize}

By comparison, prompt and controller optimization requires only two agents:

\begin{itemize}
  \item \textbf{Planning \& Coordination Agent.} Orchestrates rollout and evaluation, and manages prompt and controller versions. It analyzes image montages of recorded trajectories to identify feature estimation errors and ineffective actions, then revises prompts and controller rules accordingly.
  \item \textbf{Rollout / Capture Agent.} Executes the revised prompts and controller with fixed decision model weights, recording observations, actions, and outcomes. It returns trajectories and evaluation results to the Planning \& Coordination Agent to guide the next revision.
\end{itemize}

\FloatBarrier
\subsection{High-frequency OODA with a pipeline}
\label{sec:visual-input}

\begin{figure}[!htbp]
    \centering
    \includegraphics[width=\linewidth]{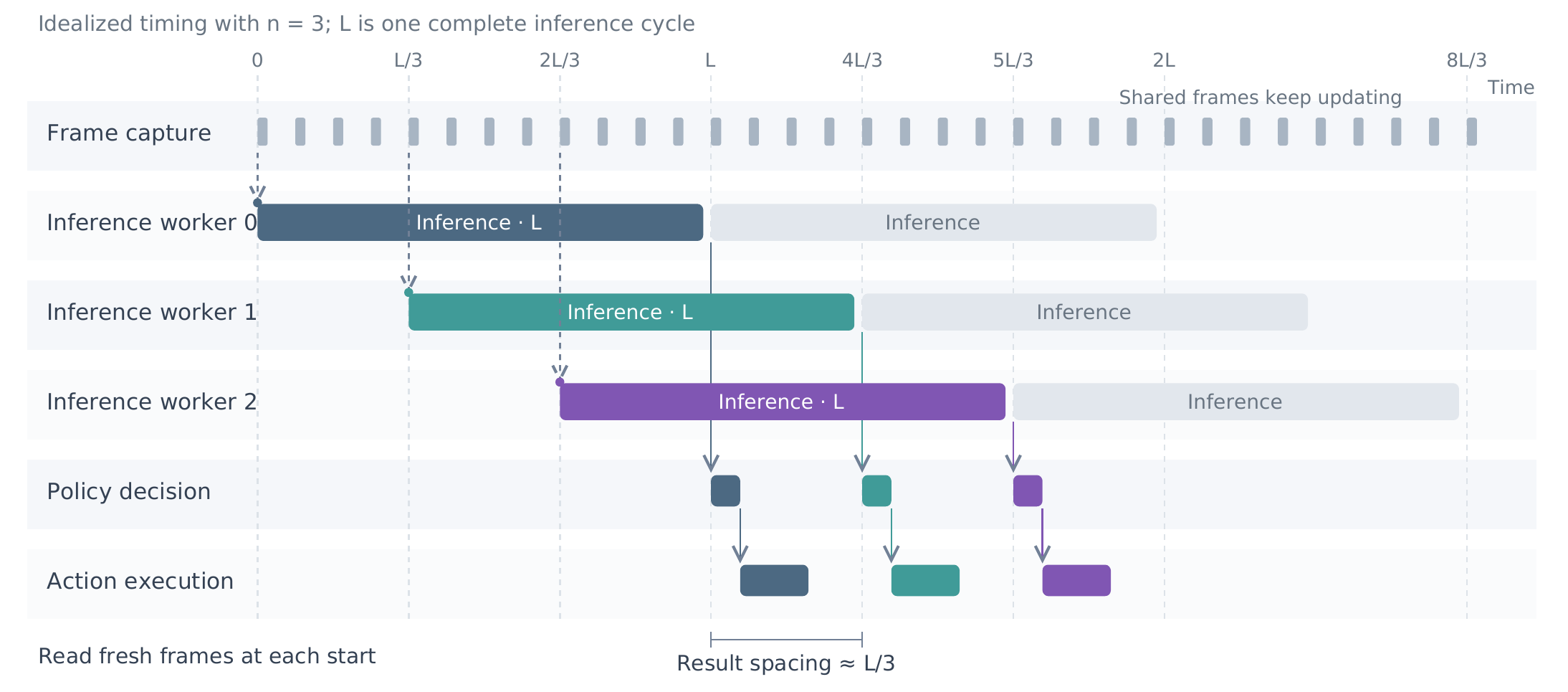}
    \caption{Idealized timing of staggered inference and independent execution.}
    \label{fig:latency-swimlanes}
\end{figure}

% Human Baseline: 由于每一帧的图像都不同，tokenizer编码后的token序列很难充分激发推理引擎（如vLLM）的kv-cache缓存加速的能力，
% 因此，决策的频率和延迟需要额外的关注。
% 在decision tree消耗的时间少到可以忽略不计，决策的延迟主要由决策模型的推理时长决定。
% 进一步地，在模型尺寸固定的情况下，耗时主要取决于输入图像的数量和分辨率，模型量化不能显著降低时延。
% 我们在实验中发现，大多数场景下，只要延迟小于一定阈值（通常约0.28秒），就足够实施高质量实时决策，
% 再降低延迟带来的收益呈现显著的边际递减。
% 反直觉的是，即使不能降低时延，我们仍然可以用流水线法，让决策频率加倍，进而提升一些环境中表现。
Because each frame in a dynamic environment differs from the previous one, the tokenizer produces a different token sequence for each frame, limiting inference engines such as vLLM in their ability to exploit key-value (KV) cache acceleration. Decision frequency and latency therefore require particular attention. The rule-based controller takes negligible time, so decision latency is determined primarily by decision-model inference. For a fixed model size, inference time depends mainly on the number and resolution of the input images; quantization cannot substantially reduce it. Experiments show that, in most scenarios, latency below a task-dependent threshold (typically about $0.28$\,s) is sufficient for high-quality real-time decisions, with markedly diminishing returns from further reductions. Even when inference latency cannot be reduced, a pipeline with $n$ concurrent inference requests can increase decision frequency by a factor of $n$ relative to serial inference and improve performance in some environments.

As shown in Fig.~\ref{fig:latency-swimlanes}, the implementation uses a pipeline that decouples millisecond-scale frame capture from the longer inference cycle. Screenshots continue to refresh the shared history $H_i$, while $n$ inference workers start at staggered intervals and each reads the latest available frames. With sufficient serving capacity, the pipeline can return results approximately $L/n$ apart, compared with one result every interval $L$ under serial inference, without shortening the latency $L$ of an individual request. The figure illustrates $n=3$ as an example; the experiments use $n=1$ for ZZZ and Snake and $n=2$ for Wukong. Because controller decisions are fast but depend on state carried across steps, a single controller processes results serially as they arrive. An independent executor runs each selected action and returns its execution timestamp while frame capture and inference continue in parallel.

\begin{figure}[!htbp]
    \centering
    \includegraphics[width=0.6\textwidth]{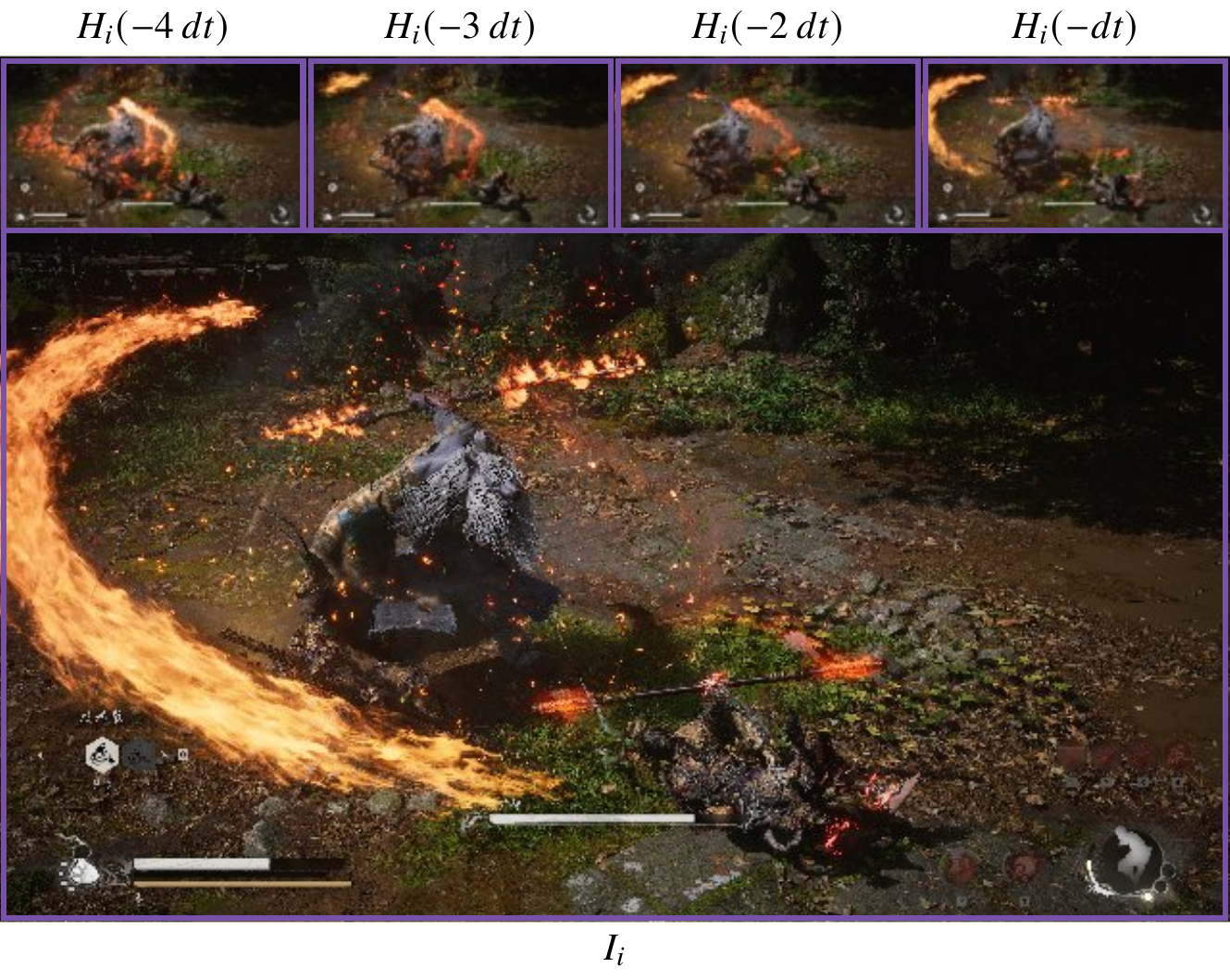}
    \caption{Four reduced-resolution history frames placed above the current observation to form the hybrid image.}
    \label{fig:cat-image}
\end{figure}

When a task requires temporal context, the latest $|H_i|$ images are retrieved from the history at the frame-capture lane in Fig.~\ref{fig:latency-swimlanes}, reduced in resolution, and placed above the current image to form a hybrid image (Fig.~\ref{fig:cat-image}). The rule-based controller then uses the resulting features to determine the action.

\FloatBarrier
\subsection{Visual decision under reaction budget}
\label{sec:model-latency}

% 段落总结：固定模型与截图，扫描分辨率，以类别均衡指标和概率损失衡量视觉输入的质量与开销。
To examine how visual-input cost constrains the computation available for a response, the resolution sweep uses 260 archived ZZZ screenshots with a fixed 27B backbone. The reference low-rank adaptation (LoRA) adapter (Base)~\cite{hu2021lora} and two fine-tuned checkpoints (\texttt{zzz-rsi-v1} and \texttt{zzz-rsi-v2}) described in Section~\ref{sec:zzz} answer the same attack-warning question. Balanced accuracy averages the recalls of all feature states, and negative log-likelihood (NLL) measures probability-prediction quality. NLL is reported both as a per-frame mean and as an equal-weight mean of the three state-specific NLLs. Each checkpoint retains its decision head and temperature throughout the sweep.

\begin{figure}[!htbp]
    \centering
    \includegraphics[width=\textwidth]{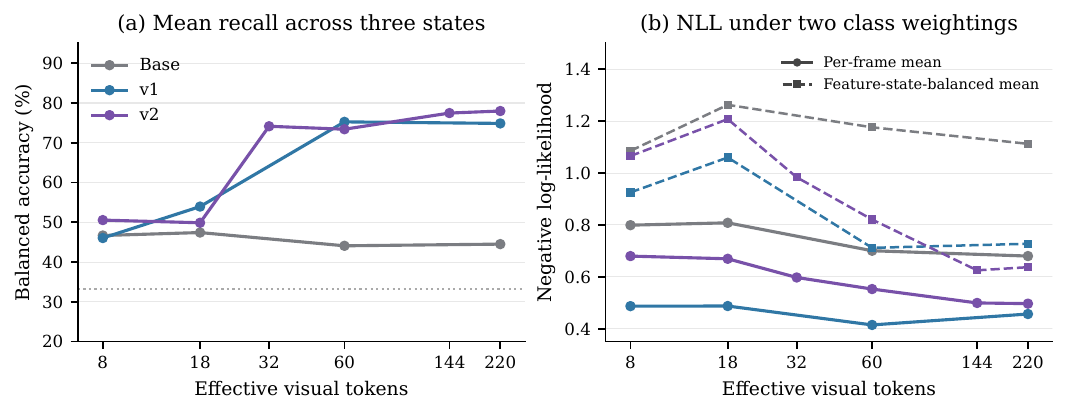}
    \caption{Resolution sensitivity on the same 260 ZZZ validation frames. Base includes the reference LoRA adapter; v1 and v2 use fine-tuned adapters on the same backbone. Balanced accuracy gives each state equal weight; the dotted line marks a constant-state predictor (33.33\%). For NLL, solid curves average over frames and dashed curves average the three state-specific means equally; colors identify checkpoints in both panels. Higher balanced accuracy and lower NLL are better. Points report the measured configurations, with lines connecting adjacent tested token counts.}
    \label{fig:zzz-resolution-quality}
\end{figure}

% 段落总结：分辨率影响并非单调，总体准确率及按帧平均的NLL可能掩盖分类损失，应联合报告均衡指标与分类结果。
Fig.~\ref{fig:zzz-resolution-quality} shows nonmonotonic sensitivity. For v2, reducing the submitted image from $640\times360$ (220 visual tokens) to $256\times144$ (32 tokens) lowers balanced accuracy from 77.98\% to 74.15\% and raises mean NLL from 0.497 to 0.597, even as top-1 accuracy rises from 81.15\% to 83.08\%. At $192\times108$ (18 tokens), balanced accuracy falls to 49.89\%. Aggregation also matters: for v1 at 18 tokens, yellow NLL increases from 0.349 to 1.693 while red NLL decreases from 1.486 to 1.131. Mean NLL changes only from 0.457 to 0.487, whereas feature-state-balanced NLL rises from 0.727 to 1.060. At 60 tokens, v1 instead has a lower mean NLL of 0.414.

\begin{figure}[!htbp]
    \centering
    \includegraphics[width=0.6\textwidth]{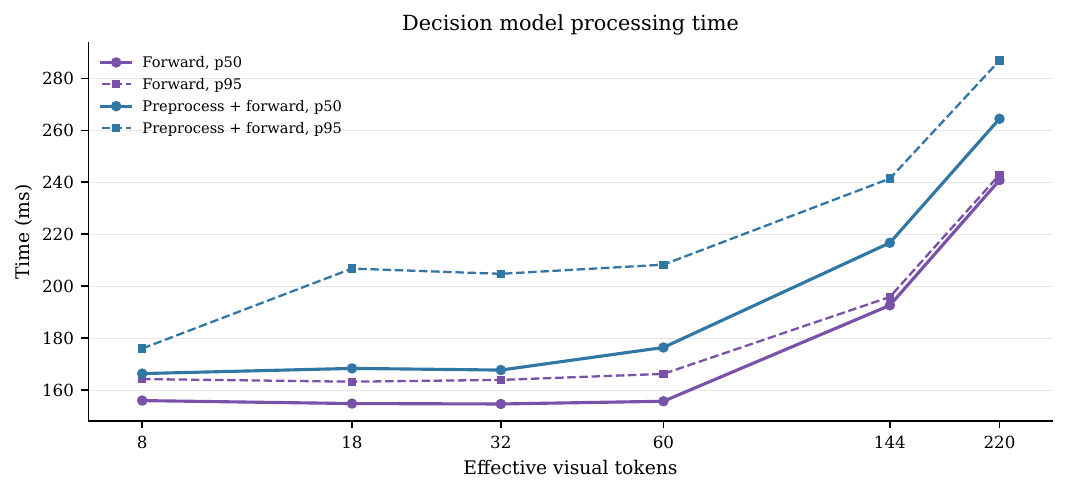}
    \caption{Decision model processing time. Timing uses the offline Transformers implementation, including reference PyTorch kernels. Preprocessing includes image decoding, resizing, processor operations, and transfer to the GPU. The processor minimum is lowered to 1{,}024 pixels to permit the resolution sweep. Percentiles are over one timed pass of 260 frames per configuration; capture, network, controller, and action execution are excluded.}
    \label{fig:zzz-resolution-timing}
\end{figure}

\FloatBarrier
% 段落总结：同步测量离线预处理和前向耗时，低 token 区间耗时趋于平台，但预警识别仍会继续退化。
Fig.~\ref{fig:zzz-resolution-timing} reports batch-size-one measurements on a single H20 GPU using BF16, with eight warm-up passes per shape and synchronized forward timing. For v2, the forward p50/p95 falls from $240.9/242.9$\,ms at 220 tokens to $154.7/164.0$\,ms at 32 tokens. Further reduction to 18 or 8 tokens leaves the median forward time near 155\,ms while warning recognition deteriorates. In this implementation, reducing the number of visual tokens eventually yields little additional saving in forward time.

\FloatBarrier

% 段落总结：本实验仅描述已有验证素材上的质量与本地计算开销，不推断独立场景泛化或端到端按时响应表现。
These measurements describe validation-set sensitivity. The ten source episodes also contribute training frames, and the validation material was used for checkpoint selection; only 11 yellow and 25 red examples are available. The observed tradeoff therefore does not establish generalization to independent episodes. Local processing time accounts for part of the reaction budget, but the archived labels provide no independent action deadlines. Establishing timely control performance, or the benefit of a smaller backbone, requires additional measurements.

\section{Experiments}
\label{sec:results}
% // AI generated
Zenless Zone Zero and Black Myth: Wukong provide visual environments in which observations and action opportunities change during execution.
Both implementations combine a 27B decision model with an OODA controller.
The ZZZ study examines decision-model fine-tuning and evaluation across opponents with a fixed controller.
The Wukong study compares policies for sustained evasion and offensive action.
A supplementary Snake study compares episode outcomes across four versions.

\subsection{Recursive improvement in Snake}
\label{sec:snake}
% // AI generated
% AI Summary: Snake 实验在固定时限的单局协议下比较基座与三个迭代版本，每个版本评估 20 局，并统计得分分布、达到 5 分的局数及可用的平均步数。
The Snake experiment compares the base configuration with three fine-tuned versions, each evaluated over 20 episodes under a fixed per-episode time limit (Table~\ref{tab:snake-summary}).
The v1 and v2 results are reevaluations of the corresponding checkpoints.
The score is the number of food items collected before the episode ends, so its absolute value is specific to this time-limited protocol; mean, median, and maximum scores summarize episode outcomes, while the number of episodes scoring at least 5 indicates how frequently each version attains this threshold.
Mean episode length is also reported where available.

\begin{figure}[!htbp]
\centering
\includegraphics[width=0.25\textwidth]{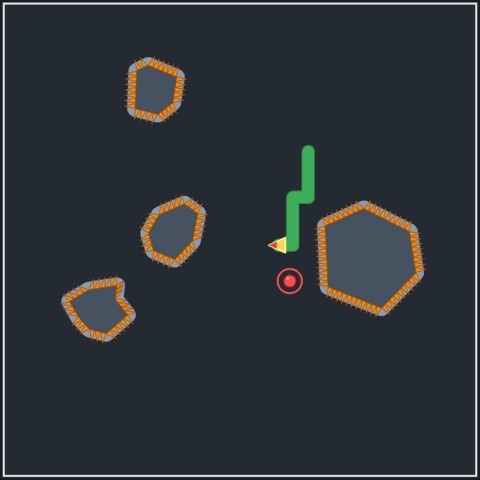}%
\includegraphics[width=0.25\textwidth]{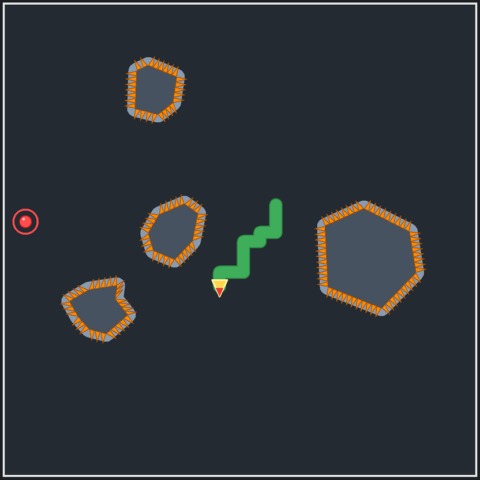}
\caption{Example observations from the Snake environment. Red dots represent food, and polygons with spikes are barriers. The environment advances at every time tick, regardless of whether an action arrives in time.}
\label{fig:snake-observations}
\end{figure}

\begin{table}[!htbp]
\centering
% // AI generated
\caption{Snake results over 20 episodes per version under a fixed per-episode time limit; v1 and v2 are reevaluations, and N/A indicates an unreported value.}
\label{tab:snake-summary}
\small
\setlength{\tabcolsep}{4pt}
\begin{tabular}{@{}lrrrrr@{}}
\toprule
Version & Mean & Median & Maximum & \shortstack{Episodes\\with score $\geq 5$} & \shortstack{Mean length\\(steps)} \\
\midrule
\texttt{snake-rsi-base} & 1.30 & 0.5 & 5 & 2/20 & N/A \\
\texttt{snake-rsi-v1} & 4.65 & 3.5 & 12 & 7/20 & 230 \\
\texttt{snake-rsi-v2} & 3.95 & 3.0 & 15 & 6/20 & 185 \\
\texttt{snake-rsi-v3} & 7.35 & 7.5 & 13 & 15/20 & N/A \\
\bottomrule
\end{tabular}
\end{table}

% // AI generated
% AI Summary: v3 的平均得分、中位数及达到 5 分的局数均优于基座，而 v2 相较 v1 的回退表明迭代收益并非单调增长。
The final version, \texttt{snake-rsi-v3}, increases the mean score from 1.30 to 7.35 and the median from 0.5 to 7.5 relative to the base configuration.
The proportion of episodes scoring at least 5 rises from 10\% to 75\%, indicating that the gain extends beyond isolated high-scoring episodes.
Improvement is not monotonic: v2 has a lower mean and median than v1 despite achieving the highest individual score of 15.
These results illustrate both the gains attainable through recursive self-improvement and the need to evaluate each recursive update across multiple episodes.

\FloatBarrier
\subsection{Visual features in Zenless Zone Zero}
\label{sec:zzz}
% // AI generated
The ZZZ controller uses attack warnings, combat state, skill readiness, and character identity to select actions.
The decision head and calibration remain unchanged.
The training and validation sets contain 922 and 164 frames for \texttt{zzz-rsi-v1} and 1{,}462 and 260 frames for \texttt{zzz-rsi-v2}, respectively.
Validation NLL is used to select each checkpoint.

\begin{figure}[!htbp]
\centering
\includegraphics[width=\textwidth]{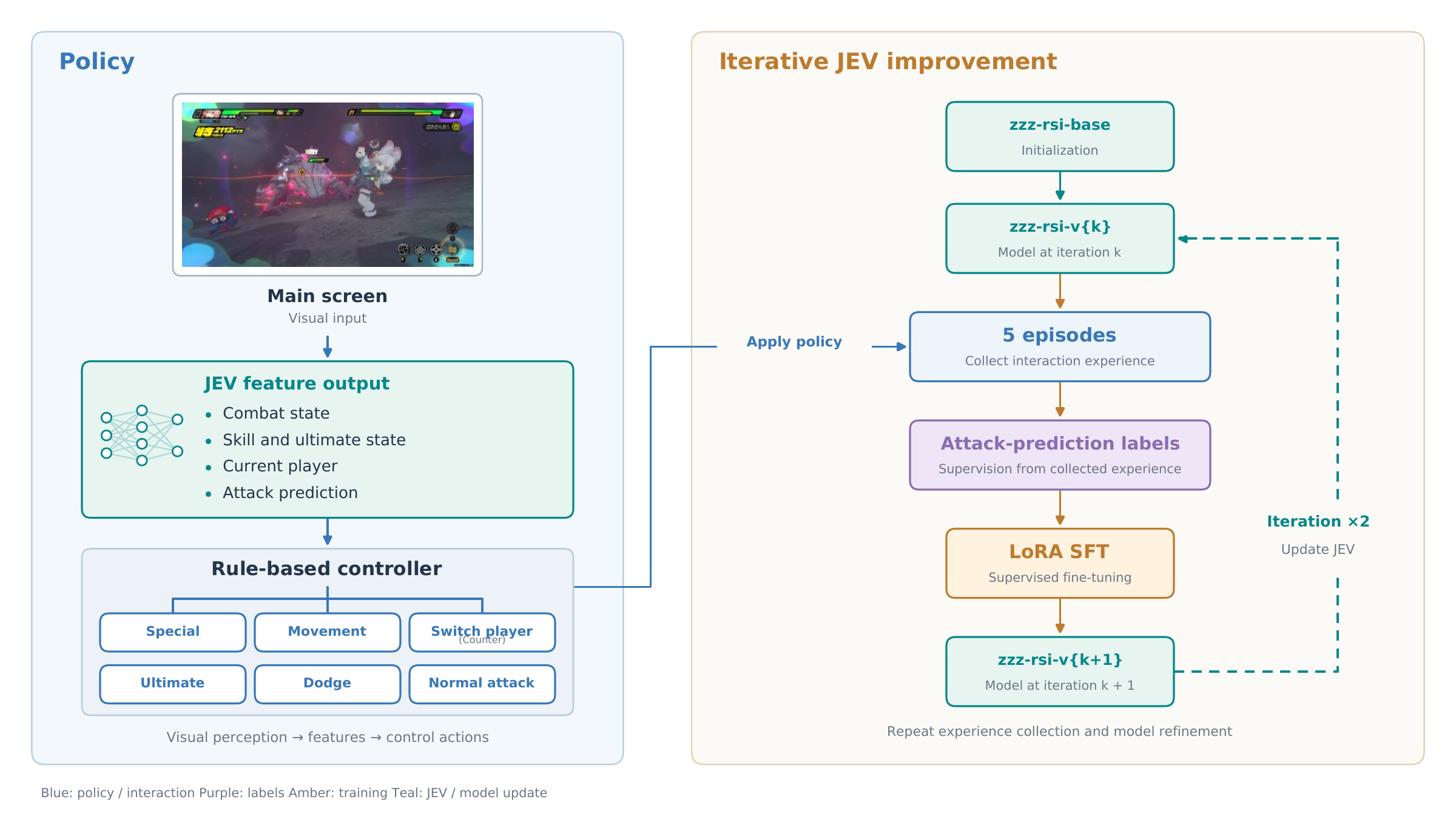}
% // AI generated
\caption{Policy execution and model fine-tuning in Zenless Zone Zero.}
\label{fig:zzz-iteration}
\end{figure}

% Evidence: data/jev260-readme.md.
For v2, the candidate pool retains every case where the logged attack-warning answer disagrees with its reference label and samples an equal number of cases where they agree. This pool is then split into training and validation sets.

% // AI generated
Fig.~\ref{fig:zzz-iteration} connects the deployed policy to the two improvement rounds.
Each round collects five episodes, labels attack warnings, and uses those examples to update the model with LoRA.
The rule-based controller converts the resulting features into skill use, movement, character switching, and attacks.

% // AI generated
The fine-tuned model returns to the same interaction process to collect the next round of experience.
The validation comparison below measures changes in the attack-warning features that drive this policy.

% // AI generated
Table~\ref{tab:validation260} shows improved feature prediction accuracy after RSI on the same 260-frame validation set. From the base model to \texttt{zzz-rsi-v2}, Top-1 accuracy increases from 63.85\% to 81.15\%, and balanced accuracy increases from 44.52\% to 77.98\%.

\begin{table}[!htbp]
\centering
% // AI generated
\caption{ZZZ attack-warning validation.}
\label{tab:validation260}
\small
\begin{tabular}{lrrr}
\toprule
Model & Top-1 (\%) & Balanced accuracy (\%) & NLL \\
\midrule
\texttt{zzz-rsi-base} & 63.85 & 44.52 & 0.6799 \\
\texttt{zzz-rsi-v1} & 79.23 & 74.87 & 0.4565 \\
\texttt{zzz-rsi-v2} & 81.15 & 77.98 & 0.4966 \\
\bottomrule
\end{tabular}
\end{table}

\FloatBarrier
\subsection{Online control in Zenless Zone Zero}
% // AI generated
Each decision-model version is evaluated over five episodes with the same two-character team and controller under a 300\,s limit. The team consists of Nekomata and Zhao, both at level 60, and is referred to as the P2 configuration.
Table~\ref{tab:zzz-online} reports completion and decision timing.
Completion duration summarizes successful episodes.
The timestamp $t_{50}$ is the first point on the episode runner's clock, measured in seconds, at which the opponent's imbalance gauge reaches at least 50\%. The table gives its median across the five episodes for each decision-model version.
The timing column gives the median and 95th percentile of logged decision-round durations from frame capture to action selection.

\begin{table}[!htbp]
\centering
% // AI generated
\caption{ZZZ outcomes and decision latency.}
\label{tab:zzz-online}
\small
\setlength{\tabcolsep}{4pt}
\begin{tabular}{lrrrr}
\toprule
Model & Wins & Median completion (s) & Median $t_{50}$ (s) & \shortstack{Decision round\\p50/p95 (ms)} \\
\midrule
\texttt{zzz-rsi-base} & 2/5 & 261.2 & 167.0 & 231.4/244.2 \\
\texttt{zzz-rsi-v1} & 5/5 & 236.2 & 26.0 & 229.1/249.4 \\
\texttt{zzz-rsi-v2} & 5/5 & 239.3 & 29.7 & 230.2/255.0 \\
\bottomrule
\end{tabular}
\end{table}

\begin{figure}[!htbp]
\centering
\includegraphics[width=\textwidth]{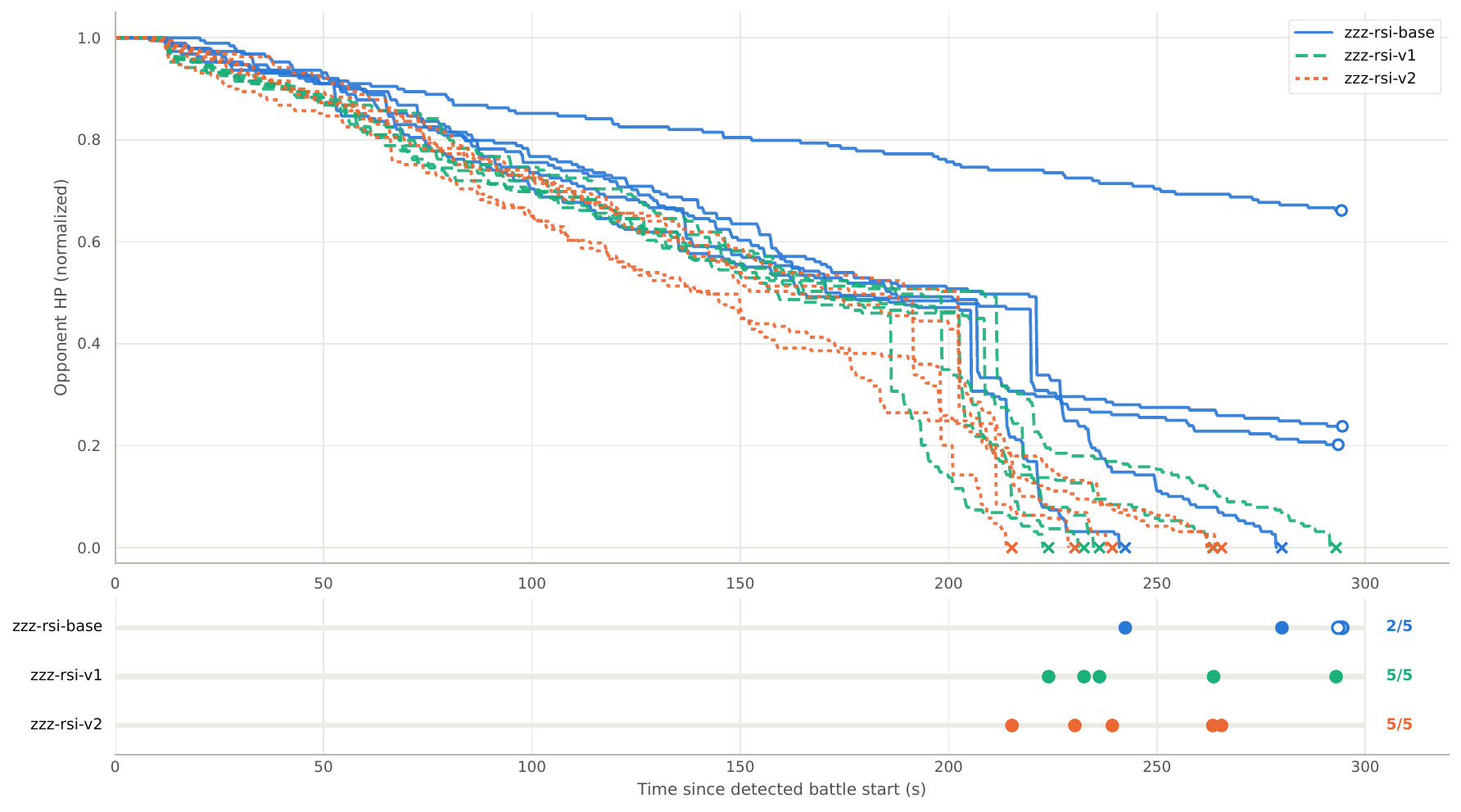}
% // AI generated
\caption{Opponent-health trajectories in fifteen ZZZ episodes.}
\label{fig:zzz-trajectories}
\end{figure}

% // AI generated
Runs with either fine-tuned decision model complete all five episodes, compared with two completions for the reference model.
The median decision round lasts approximately 230\,ms with each decision-model version.
Fig.~\ref{fig:zzz-trajectories} shows how opponent health changes during each episode.
Crosses mark successful completion, and open circles mark unfinished episodes at the end of the record.
The lower panel places the outcomes on the same time axis.

% // AI generated
The trajectories show differences in the pace of combat even when the final outcomes agree.
With five unpaired episodes per decision-model version, they do not establish an ordering between \texttt{zzz-rsi-v1} and \texttt{zzz-rsi-v2}.
The accompanying \href{https://public.agent-matrix.com/publish/auto/jev-v2/ooda-base-ep4-310-425-20261101.mp4}{reference} and \href{https://public.agent-matrix.com/publish/auto/jev-v2/ooda-v1-ep3-2m-20261101.mp4}{fine-tuned-model recordings} show how visual features enter the action rules.

\FloatBarrier
\subsection{Transfer OODA policy from YeShiyuan to other ZZZ levels}
% // AI generated
The ZZZ controller used for YeShiyuan is also evaluated on nine additional opponents under the P2 team configuration.
The controller rules remain fixed across levels and decision-model versions.
These evaluations comprise 39 attempts, of which 26 complete the task.
Here, \texttt{zzz-rsi-v2} and \texttt{zzz-rsi-v3} identify the decision-model versions used by the shared controller.

\begin{figure}[!htbp]
\centering
\includegraphics[width=\textwidth]{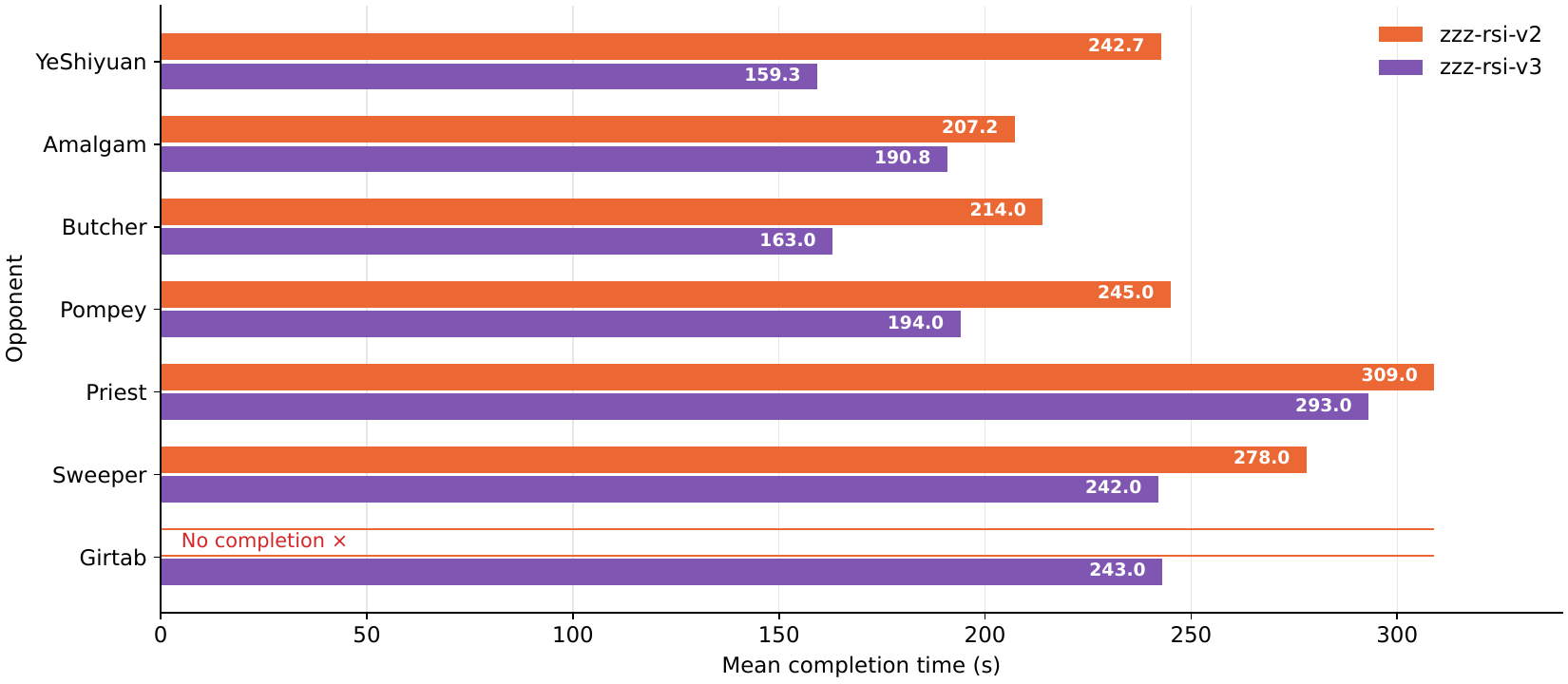}
% // AI generated
\caption{Transfer OODA policy from YeShiyuan to other ZZZ levels with a shared controller. Compared with \texttt{zzz-rsi-v2}, \texttt{zzz-rsi-v3} achieves shorter mean completion times on levels completed by both versions and completes Girtab, where v2 has no completed attempts. Bars show mean times over completed episodes only.}
\label{fig:controller-means}
\end{figure}

% // AI generated
Fig.~\ref{fig:controller-means} compares mean completion times for opponents with at least one completed episode.
Numbers inside the bars give the mean completion times in seconds.
YeShiyuan is a separate comparison with five \texttt{zzz-rsi-v2} episodes and three \texttt{zzz-rsi-v3} episodes, outside the nine-opponent attempt count.
All means condition on completion, so an unfinished attempt is not assigned an artificial completion time.

% // AI generated
For Amalgam, runs with each decision-model version complete five episodes, with mean time decreasing by 16.4\,s from 207.2 to 190.8\,s.
With \texttt{zzz-rsi-v3}, the controller completes three YeShiyuan episodes in 162, 165, and 151\,s, averaging 159.3\,s.
For Priest, completed episodes take 309 and 293\,s with \texttt{zzz-rsi-v2} and \texttt{zzz-rsi-v3}, respectively; each version also has one unfinished attempt.
With \texttt{zzz-rsi-v2}, the controller does not complete either Girtab attempt; with \texttt{zzz-rsi-v3}, it completes two in 257 and 229\,s.
Hunter, Twins, and Bringer have no completed episodes in this evaluation.
Sample counts and completion subsets differ between opponents, so these means describe the observed runs rather than a uniform improvement rate.

\FloatBarrier
\subsection{Evasive control in Black Myth: Wukong}
\label{sec:wukong}

% // AI generated
Fig.~\ref{fig:prompt-case} shows why temporal features require more than a coarse description of posture.
The second and third observations both show a raised weapon but occur at different stages before the attack.

% // AI generated
Distinguishing panels (b) and (c) requires attention to weapon angle and the recent change in posture.
A short description can merge these states, while a longer description must still identify the visual features that determine action timing.
Learning from observation sequences provides a mechanism for encoding such distinctions in parameters rather than repeatedly describing them in the prompt.

% // AI generated
The Wukong model predicts the remaining time before damage from the current observation and four historical frames.
A prediction inside the configured dodge window can trigger a dodge once the cooldown has cleared.

\begin{figure}[!htbp]
\centering
\includegraphics[width=\textwidth]{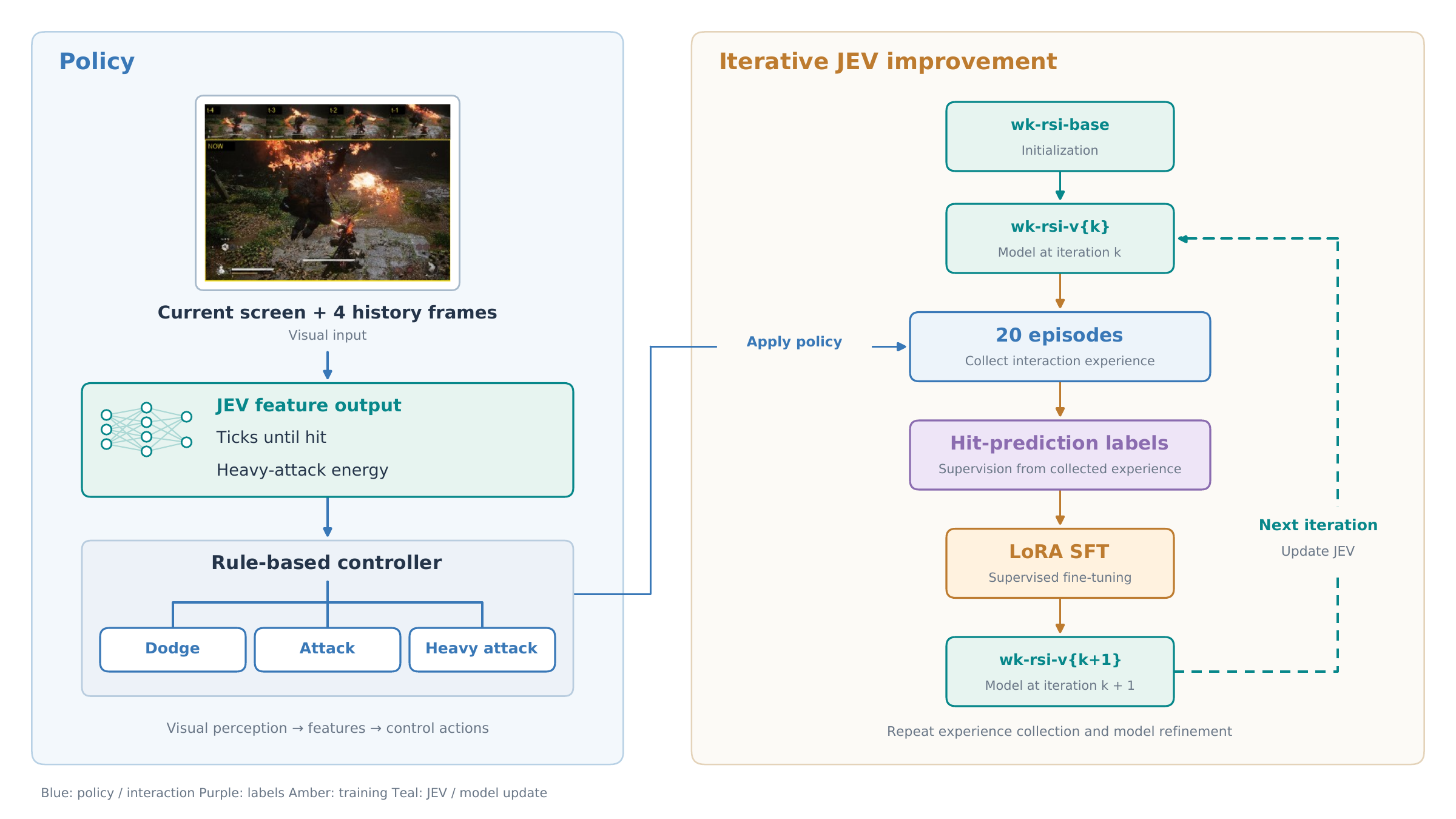}
% // AI generated
\caption{Policy execution and model fine-tuning in Black Myth: Wukong.}
\label{fig:wukong-iteration}
\end{figure}
\FloatBarrier

% Evidence: docs/evidence/wukong_survival_20261104/source/jev_prompts.py and README.
% The archived sources disagree about the precise tick rounding rule.
In the archived v1/v2 runs, a qualifying hit is identified by a visible loss of player health. Labels use the time from frame capture to the next hit, expressed on a $180$\,ms tick scale with answers from $0$ through $5$. A fully observed prediction window with no qualifying hit is labeled \emph{no immediate damage}. The labels describe outcomes under the actions recorded in the trajectory.

% Evidence: docs/evidence/wukong_survival_20261104/training/v1_v2_training_meta.md.
The v2 data preparation adds all new examples with a numbered hit label and a sample of the no-immediate-damage examples to the preceding training set. The v2 fit starts from the v1 adapter. Both fits use the class-weighted focal loss in Eq.~\ref{eq:focal-loss} with $\gamma=2$~\cite{lin2017focal}.

Fig.~\ref{fig:wukong-iteration} shows how recent observations, predicted time to damage, and interaction data enter the recursive improvement process.
Its action branches describe the broader policy family; the evasive evaluation below enables only the evasive actions.

% // AI generated
The illustrated training cycle collects interactions, constructs hit-prediction labels, and updates the model before redeployment.
The evasive policy does not issue attacks, heavy attacks, or healing actions.
Table~\ref{tab:wukong-evasion} compares recorded episode durations with a stationary reference that takes no active actions.

% Keep the duration plot with its introduction when earlier content changes.
\par\noindent\begin{minipage}{\linewidth}
Fig.~\ref{fig:wukong-duration} plots the group means and medians, with episode counts alongside each row.
Vertical marks denote medians; the stationary reference has no median reported in this comparison.

\begin{figure}[H]
\centering
\includegraphics[width=\textwidth]{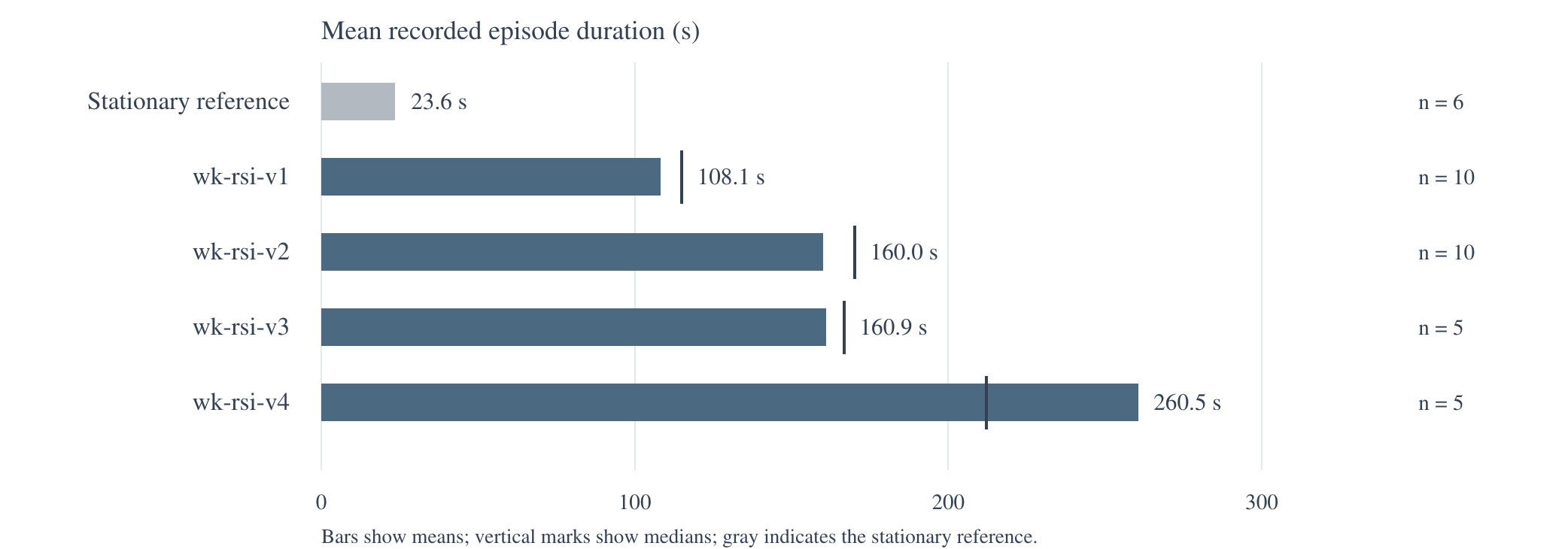}
% // AI generated
\caption{Recorded durations under the Wukong evasive policy.}
\label{fig:wukong-duration}
\end{figure}
\end{minipage}\par
\FloatBarrier

% // AI generated
The gap between the mean and median summarizes differences within each set of recorded episodes.
Table~\ref{tab:wukong-evasion} provides the numerical values and separates the experimental batches.

\begin{table}[!htbp]
\centering
% // AI generated
\caption{Wukong evasive-policy durations.}
\label{tab:wukong-evasion}
\small
\begin{tabular}{lrrr}
\toprule
Model & Episodes & Mean (s) & Median (s) \\
\midrule
Stationary reference & 6 & 23.6 & N/A \\
\midrule
\texttt{wk-rsi-v1} & 10 & 108.1 & 114.9 \\
\texttt{wk-rsi-v2} & 10 & 160.0 & 170.1 \\
\texttt{wk-rsi-v3} & 5 & 160.9 & 166.7 \\
\texttt{wk-rsi-v4} & 5 & 260.5 & 212.2 \\
\bottomrule
\end{tabular}
\end{table}

\begin{figure}[!htbp]
\centering
\setlength{\tabcolsep}{2pt}
\begin{tabular}{@{}cc@{}}
\includegraphics[width=.49\textwidth]{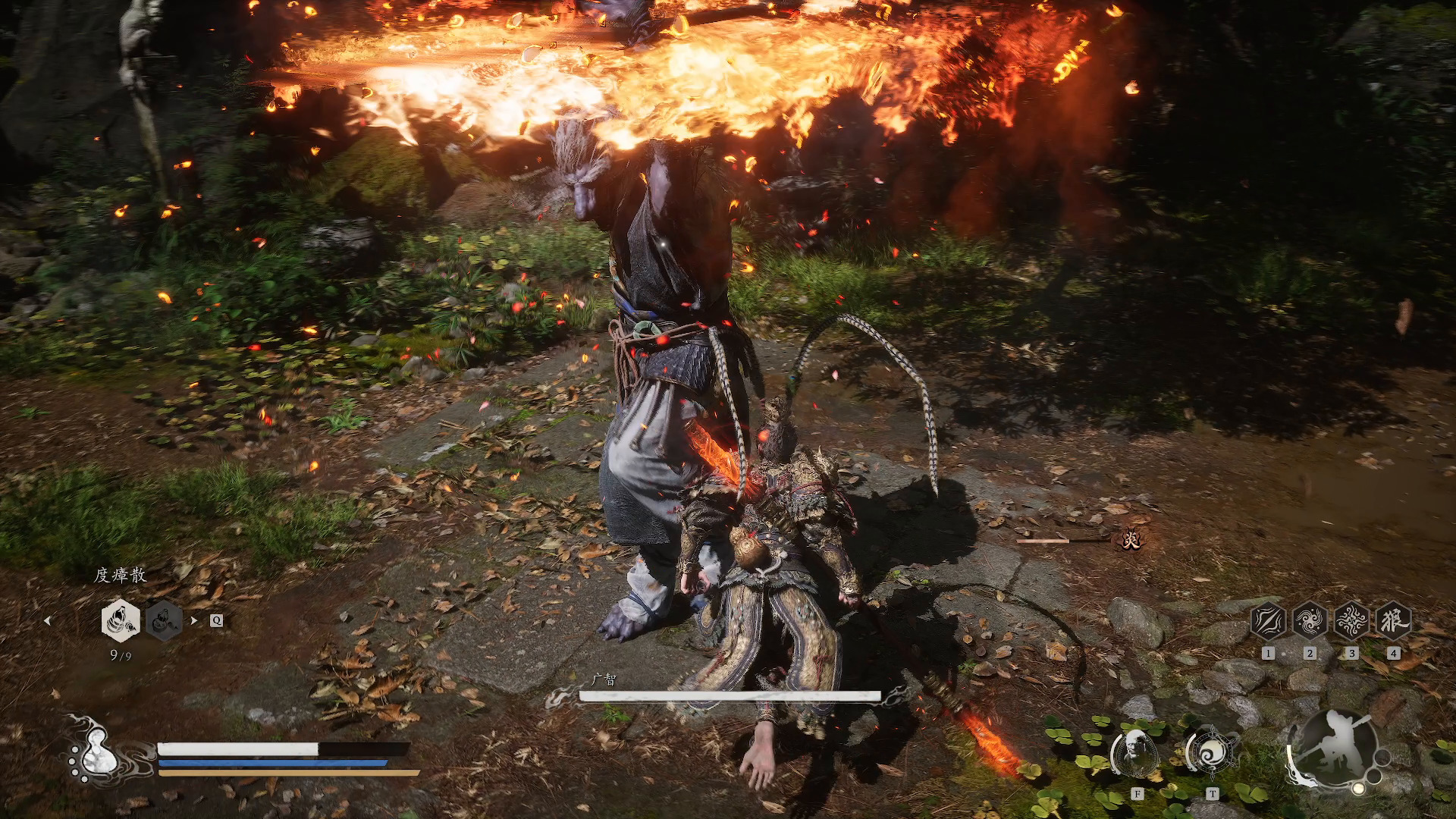} &
\includegraphics[width=.49\textwidth]{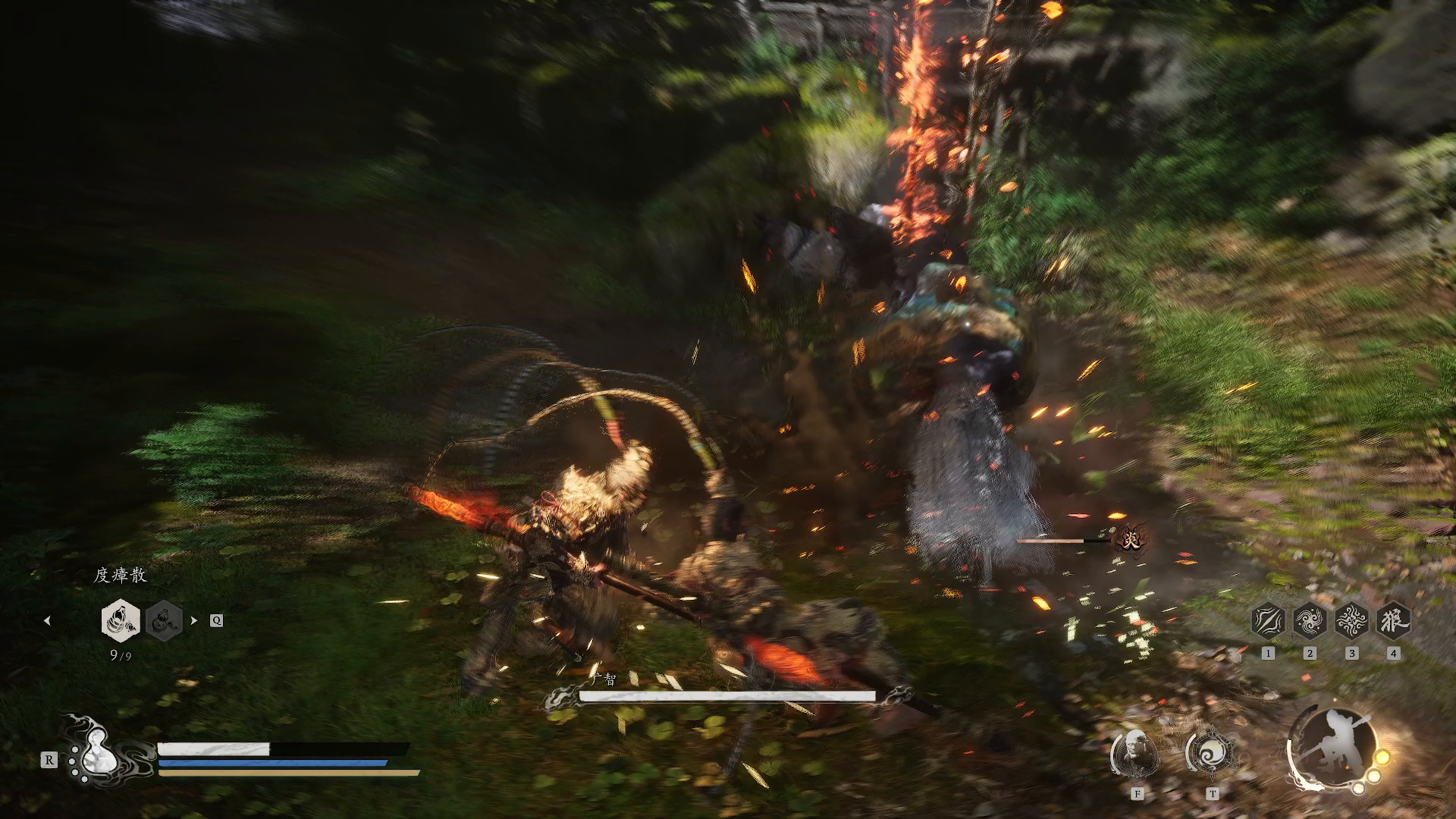} \\
(a) & (b)
\end{tabular}
% // AI generated
\caption{Visual observations during evasive control in Black Myth: Wukong.}
\label{fig:wukong-evasion}
\end{figure}

% // AI generated
Under the later configuration, mean duration increases from 160.9\,s for \texttt{wk-rsi-v3} to 260.5\,s for \texttt{wk-rsi-v4}, a difference of approximately 62\%.
The earlier v1/v2 batches use a 350\,ms action cooldown, while the later v3/v4 batches use 500\,ms.
Recorded durations include the time required to detect termination and do not identify an exact time of death.
The batches therefore remain separate comparisons.
Fig.~\ref{fig:wukong-evasion} shows the changing visual conditions in which these responses are selected.

% // AI generated
Both observations contain rapid motion and occlusion near the opponent.
Recent frames provide information about how the action has developed, while the current frame constrains the timing of the response.
Task completion also requires choosing when to attack, motivating a separate offensive policy.

\FloatBarrier
\subsection{Offensive control in Black Myth: Wukong}
% // AI generated
The offensive policy coordinates dodging, attacks, and healing.
Win rate measures task completion, and mean winning-episode duration describes the time taken by successful attempts.
Table~\ref{tab:wukong-offense} reports the five model configurations in two experimental batches.
The first three configurations each run ten episodes; the remaining two each run fifteen.
The first three configurations use the same offensive controller, allowing a within-batch comparison across JEV checkpoints.

\begin{table}[!htbp]
\centering
% // AI generated
\caption{Wukong offensive-policy outcomes.}
\label{tab:wukong-offense}
\small
\setlength{\tabcolsep}{4pt}
\begin{tabular}{lrrrr}
\toprule
Model & Episodes & Win rate (\%) & Mean duration (s) & Winning mean (s) \\
\midrule
\texttt{wk-rsi-base} & 10 & 0.0 & 21.9 & N/A \\
\texttt{wk-rsi-v1} & 10 & 20.0 & 74.1 & 139.7 \\
\texttt{wk-rsi-v2} & 10 & 30.0 & 89.5 & 122.4 \\
\texttt{wk-rsi-v3} & 15 & 60.0 & 139.7 & 154 \\
\texttt{wk-rsi-v4} & 15 & 66.7 & 125.3 & 131 \\
\bottomrule
\end{tabular}
\end{table}

% // AI generated
The base model wins none of its ten episodes, while every updated configuration has at least one win.
The \texttt{wk-rsi-v4} configuration wins ten of fifteen episodes, with a mean winning-episode duration of 131\,s.
The two batches use different controller settings, and successful episodes form different subsets of the attempted runs.
The values do not isolate the effect of model fine-tuning across all five configurations.
The \href{https://public.agent-matrix.com/publish/auto/wukong-videos/wukong-v2-best-ep.mp4}{evasive-policy recording} and \href{https://public.agent-matrix.com/publish/auto/wukong-videos/win2case-trim-small.mp4}{offensive-policy recording} illustrate the action timing behind these aggregate outcomes.

\section{Discussion}
\label{sec:discussion}
\label{sec:limitations}

System-One RSI separates reasoning about a task from responding to it during execution.
The reasoning agents use recorded interactions to guide learning and revise the policy, while the online loop uses the resulting predictions and rules to act.
This division shapes both the practical advantages of the framework and the conditions under which it can be applied.

\subsection*{Advantages}
System-One RSI builds muscle memory for real-time agents through reasoning-guided few-shot practice.
Reasoning agents guide learning from a small number of interaction episodes, and the decision model retains the learned distinctions for rapid use during execution.
This approach offers the following advantages:
\begin{itemize}
% Human - AI Assisted
% Human Baseline: 少量真实交互即可学习。面向无法沙盒化、无法并行运行的非仿真环境，通过少量实际交互形成任务能力。
% AI Summary: 系统通过少量真实交互学习任务能力，适用于无法沙盒化或并行运行的非仿真环境。
\item \textbf{Learning from limited real interaction.}
The framework develops task capabilities through limited interaction with live environments that cannot be sandboxed or run in parallel.

% Human - AI Assisted
% Human Baseline: 保持低延迟的决策回路。推理与学习的成果进入 System One 模型，在线调用即可快速获得判断。OODA 回路据此持续执行动作。
% AI Summary: 推理与学习的成果固化到 System One 模型中，为 OODA 回路提供低延迟判断以持续执行动作。
\item \textbf{Maintaining a low-latency decision loop.}
The outcomes of reasoning and learning are consolidated in the System-One model, which supplies rapid features during online inference and enables the OODA loop to execute actions continuously.

% Human - AI Assisted
% Human Baseline: 利用先验知识构建初始策略。网页搜索和人类描述提供任务知识。少量探索性交互补充环境反馈，用于修订初始策略。
% AI Summary: 网页搜索与人类描述提供初始任务知识，少量探索性交互提供反馈以修订初始策略。
\item \textbf{Using prior knowledge to construct initial policies.}
Web search and human descriptions provide task knowledge, while limited exploratory interaction supplies environmental feedback for revising the initial policy.

% Human - AI Assisted
% Human Baseline: 支持多种优化模式。推理模型帮助定位性能瓶颈。系统可固定 JEV 参数，集中修改行为树；也可固定行为树，训练 JEV。
% AI Summary: 推理模型帮助定位瓶颈，支持固定 JEV 修改行为树或固定行为树训练 JEV。
\item \textbf{Supporting multiple improvement modes.}
Reasoning models help identify performance bottlenecks, allowing the system to revise the rule-based controller while holding JEV parameters fixed, or train JEV while keeping the rule-based controller unchanged.

% Human - AI Assisted
% Human Baseline: OODA 策略便于修改和迁移。行为树明确记录判断与动作之间的关系。角色切换等任务变化通常只需调整部分规则，不必重新训练模型。
% AI Summary: 行为树明确判断与动作的对应关系，角色切换等任务变化通常可通过局部规则调整实现而无需重训模型。
\item \textbf{Facilitating OODA policy revision and transfer.}
Rule-based controllers explicitly encode the relationships between features and actions, so task changes such as character switching typically require adjusting only a subset of rules without retraining the model.

% Human - AI Assisted
% Human Baseline: JEV 可配合不同任务调整判断内容。推理模型增删提示词，修改需要判断的状态。结合行为树调整，适配后的 JEV 可用于更复杂的策略。
% AI Summary: 推理模型通过增删提示词调整 JEV 的判断目标，并结合行为树修改支持更复杂的策略。
\item \textbf{Configuring JEV features for different tasks.}
Reasoning models add or remove prompts to change the states that JEV evaluates; together with rule-based controller revisions, these changes allow a fine-tuned JEV model to support more complex policies.

% Human - AI Assisted
% Human Baseline: 无需专家操作示范。训练样本来自环境交互，不采集人类操作数据。公开文本与图像仍用于提供任务知识。
% AI Summary: 训练数据来自环境交互而非人类操作示范，公开文本与图像仍提供任务知识。
\item \textbf{Learning without expert operational demonstrations.}
Training examples come from environment interaction without collecting human operation data, while public text and images remain sources of task knowledge.
\end{itemize}

These advantages connect the two loops: reasoning guides what to learn or revise, and the updated online policy produces the experience for the next round.
The ability to update model parameters and explicit rules separately also allows different sources of failure to be addressed within the same workflow.

\subsection*{Limitations}
The same design depends on the knowledge used to initialize the policy, the experience available for adaptation, and the resources supporting execution.
These dependencies impose several limits on transfer and deployment:
\begin{itemize}
% Human - AI Assisted
% Human Baseline: 跨场景迁移仍需适配。本文实验用少量交互样本解决特定任务，不采集专家操作示范。完成 2–3 次迭代所需的数据，可在约 1–10 小时内采集。形成的策略仍依赖当前环境。迁移时可能需要修改提示词与行为树，并采集新数据。
% AI Summary: 少量交互可支持特定任务的递归改进，但跨场景迁移仍可能需要修改提示词与行为树并采集新数据。
\item \textbf{Task-specific transfer.}
The experiments address specific tasks using limited interaction data without collecting expert operational demonstrations.
The data required for two to three rounds of recursive improvement can be collected in approximately 1--10 hours.
The resulting policies remain dependent on the current environment, and transfer may require revised prompts and rule-based controllers together with newly collected data.

% Human - AI Assisted
% Human Baseline: 初始策略依赖先验知识。不采集专家操作示范，并不意味着不需要人类知识。公开材料与人类描述提供任务背景，如能量机制和角色技能。这些知识帮助确定观测目标与可选动作。先验缺失或不准确时，策略构建会更困难。
% AI Summary: 初始策略依赖公开材料与人类描述提供的任务先验，先验缺失或不准确会增加策略构建难度。
\item \textbf{Dependence on prior knowledge.}
Avoiding expert operational demonstrations does not eliminate the need for human knowledge.
Public materials and human descriptions provide task context, including energy mechanics and character abilities, which helps identify relevant observations and available actions.
Missing or inaccurate prior knowledge makes policy construction more difficult.

% Human - AI Assisted
% Human Baseline: 低延迟运行需要相应硬件。当前部署使用至少两块 GPU，分别承担环境渲染与 JEV 推理，减少资源竞争。这增加了硬件成本，也限制了单 GPU 设备上的直接部署。
% AI Summary: 当前部署至少使用两块 GPU 分担环境渲染与 JEV 推理，以降低资源竞争，但增加硬件成本并限制单 GPU 直接部署。
\item \textbf{Hardware requirements for low-latency execution.}
The current deployment uses at least two GPUs, with environment rendering and JEV inference assigned to separate devices to reduce resource contention.
This arrangement increases hardware costs and limits direct deployment on single-GPU devices.
\end{itemize}

Applying the framework to a new setting therefore requires checking whether its existing features and rules remain useful and whether the available hardware meets the task's timing requirements.
When these conditions change, further guided interaction may be needed to revise the policy before reliable online execution is possible.

\section{Conclusion}
\label{sec:conclusion}
% // AI generated
\rec{} connects task understanding to executable capability through targeted practice under informal verbal guidance.
A team of reasoning-model agents supplies task knowledge and controller rules and coordinates experience collection, supervision, and training; the decision model retains learned perceptual features in its parameters as external muscle memory for online OODA execution.
The separation between these components supports targeted model updates, controller revision, and reuse when the relevant task assumptions remain valid.
The framework addresses capability formation in dynamic environments where expert demonstrations and parallel simulation may be unavailable.
The ZZZ and Wukong deployments illustrate the connection between learned features and online action, while small samples and different experimental settings limit comparative conclusions.
Its effectiveness depends on the quality of prior knowledge and supervision, the coverage of collected experience, and the resources available for deployment.

\end{document}